%% file: main.tex
\pdfoutput=1

\documentclass[11pt]{article}

\input{preamble}

\begin{document}
\maketitle

\begin{abstract}
Supervised learning relies on data annotation which usually is time-consuming and therefore expensive. A longstanding strategy to reduce annotation costs is {\em active learning}, an iterative process, in which a human annotates only data instances deemed informative by a model. Research in active learning has made considerable progress, especially with the rise of large language models (LLMs). However, we still know little about how these remarkable advances have translated into real-world applications, or contributed to removing key barriers to active learning adoption. To fill in this gap, we conduct {\em an online survey in the NLP community} to collect previously intangible insights on current implementation practices, common obstacles in application, and future prospects in active learning. We also reassess the perceived relevance of data annotation and active learning as fundamental assumptions. Our findings show that data annotation is expected to remain important and active learning to stay relevant while benefiting from LLMs. Consistent with a community survey from over 15 years ago, three key challenges yet persist---setup complexity, uncertain cost reduction, and tooling---for which we propose alleviation strategies. We publish an anonymized version of the dataset.\customfootnote{{}$^\ast$Equal contribution.}\customfootnote{\hspace{-2pt}{}$^\clubsuit$Corresponding author: \href{mailto:Julia.Romberg@gesis.org}{Julia.Romberg@gesis.org}}\footnote{Dataset is available from: \href{https://doi.org/10.7802/2990}{https://doi.org/10.7802/2990}.}
\end{abstract}

\input{part1-introduction}
\input{part2-survey}
\input{part3-results}
\input{part4-related-work}
\input{part5-discussion}
\input{part6-conclusions}
\input{part-limitations}
\input{part-ethics}
\input{part-acknowledgments}

\bibliography{custom}
\input{part-appendix}

\end{document}

%% file: part1-introduction.tex
\section{Introduction}
\label{sec:introduction}

\begin{figure}[t]
\includegraphics[width=0.485\textwidth]{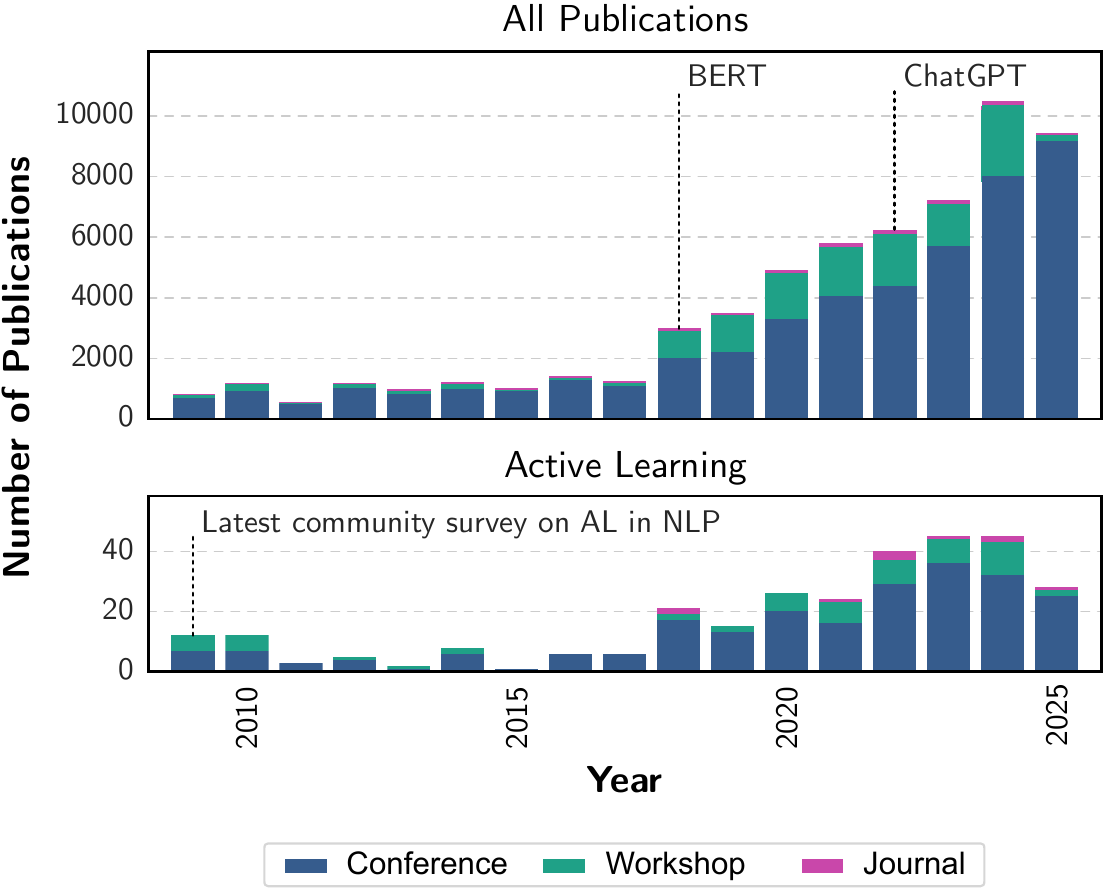}
\caption{Development of AL publication numbers in *CL venues over the past 15 years (total publication numbers for reference). The methodological details are provided in Appendix~\ref{sec:app:publication-identification}. Note that the numbers for 2025 must be interpreted with caution, as not all data were  available in the ACL Anthology at the time of analysis.}
\label{conference-publications-versus-active-learning}
\end{figure}

Supervised learning is one of the most common concepts in natural language processing (NLP). By definition, the approach depends on annotated data, which is usually time-consuming to create and therefore expensive, making supervised learning a resource-intensive process.

A longstanding strategy for minimizing the annotation effort while maintaining model performance is {\em active learning} (AL; \citealp{lewis:1994}), where a human annotator iteratively provides labels for small batches of data instances deemed informative by a model. The selection of these instances is guided by specific query strategies. After each round, the model is retrained on the newly annotated data to improve task performance. This process continues until a predetermined stopping criterion is met, concluding the annotation process.

Since its beginnings, AL has evolved considerably~\cite{xia-etal-2025-selection}, not least because of the gains brought about by large language models (LLMs\footnote{We adopt the definition of LLMs from \citet{rogers2024position}, which encompasses both encoders and decoders.}). We observe notable surges in the number of AL publications in *CL venues in 2018 and 2022 (see Figure~\ref{conference-publications-versus-active-learning}), likely due to new opportunities enabled by transformer-based language models and generative AI. These works have found considerable improvements in effectiveness and sample efficiency when integrating LLMs with AL in different scenarios (e.g., \citealp{ein-dor:2020,margatina:2021,tonneau:2022,xiao:2023}).

While AL is continuously developed in research, its primary goal remains to enable practical applications. When \citeauthor{tomanek-olsson-2009-web} last surveyed NLP practitioners on the use of AL in 2009, only 20\% stated they had implemented AL to support annotation---although many were aware of AL and its experimentally proven promises. Key barriers to adoption were skepticism about the practical effectiveness and the overhead of implementing annotation interfaces. The few that had adopted AL highlighted sampling complexity, annotator waiting times, and interface design as issues critical to the practical realization. Little do we, as a field, know about how the practical use of AL has altered since then, or whether research advancements have helped address its practical challenges. More than 15 years later, the question of practical adoption thus deserves renewed attention.

Drawing from the literature alone cannot provide a complete answer to this question, as practical use of AL---in contrast to experimental evaluations---is often not documented publicly. Therefore, we devise and conduct an extensive community survey following the example of \citet{tomanek-olsson-2009-web}. While still uncommon in NLP, community surveys have proven effective for obtaining knowledge unavailable in academic literature (e.g., \citealp{zhou:2022, subramonian:2023, michael:2023, blaschke:2024}). This way, we can not only assess the assumed transformative impact of LLMs, but also take a step back to re-examine whether fundamental assumptions about the need for data-efficient annotation still hold.

\paragraph{Contributions}
(1)~We collect a dataset of insights on AL and data annotation needs from the NLP community, academia and industry, through a comprehensive survey with 52 questions. 
(2)~We evaluate the responses of \surveyNumParticipants{} participants quantitatively and qualitatively, and compare them to the community survey from 2009.
(3)~We discuss the results and connect them to literature, describing anticipated future developments in applied AL.
(4)~We identify three key problems to adoption that persist in contemporary AL and outline possible solutions.

\paragraph{Findings} 
Our descriptive analysis indicates that data annotation remains important and a bottleneck for supervised learning, for which AL is mostly regarded as a relevant method to address this issue. Advances in NLP have led to LLMs becoming the prevalent choice of backbone model in practice, but key barriers to AL adoption from over a decade ago still persist, demanding greater emphasis on reducing the complexity of setup, ensuring cost reduction, and improving annotation tools.

%% file: part2-survey.tex
\section{Survey Methodology}

To fill the knowledge gap between LLM-driven achievements in AL literature and practical implementations, our community survey centers on five research questions. The first two assess assumptions fundamental to AL about the continued relevance of data annotation and AL as a method.

\begin{enumerate}[leftmargin=1cm]
\setlength{\itemsep}{0pt}
\item[RQ1.] What is the current state of data annotation needs in the age of LLMs?
\item[RQ2.] How does AL compare to other methods for overcoming data annotation challenges?
\end{enumerate}

\begin{figure}[!t]
\includegraphics[width=0.485\textwidth]{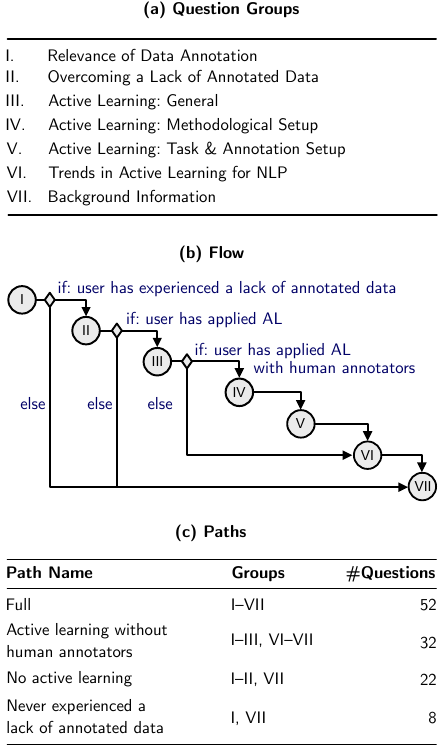}
\caption{This diagram illustrates (a)~the names and sequence of question groups in the survey, (b)~the logical flow between these question groups, including branching points and conditions, and (c) the possible paths that respondents take through the survey, ordered by the number of questions in descending order.}
\label{survey-structure-and-paths}
\end{figure}

\noindent We then turn to the main focus of interest, the contemporary implementation of AL with respect to the past \cite{olsson2009survey} and anticipated future.

\begin{enumerate}[leftmargin=1cm]
\setlength{\itemsep}{0pt}
\item[RQ3.] What do contemporary AL implementations look like in practice?
\item[RQ4.] What developments in AL have been observed and are anticipated?
\item[RQ5.] How do current perspectives on AL compare with those from 15 years ago?
\end{enumerate}

\subsection{Descriptive Research Approach}

Following previous community surveys in *CL venues~\cite{zhou:2022,michael:2023, blaschke:2024}, this survey adopts a descriptive research design in order to systematically document experiences within a sample of NLP practitioners. We capture firsthand experiences from academic and industry practitioners relevant to answering the five research questions, including real-world practices, barriers, and further perceptions of AL adoption. Our primary goal with this approach is to understand how and why practitioners use (or avoid) AL in their specific contexts, instead of making universal claims about AL's effectiveness. Notably, descriptive research methodology provides valuable documentation, even if it does not generalize to the entire target population. The practitioners' responses provide empirical evidence that has at most been theoretically discussed, but not systematically documented before. Our findings can be interpreted as valuable documentation of how AL is actually employed in practice, with important implications for future research.

\subsection{Questions and Branching} 

Our survey comprises \surveyNumQuestions{}~questions over 7~logical groups~(Figure~\ref{survey-structure-and-paths}a), which we will refer to by question group I to VII in the following. The survey includes both predefined and free-form response options. We refer to the latter as {\em other} responses, and assess them through qualitative coding, involving normalization and grouping of the inputs.

We use branching logic (Figure~\ref{survey-structure-and-paths}b) to guide participants through the survey. The main criteria here are having (1)~experienced a lack of annotated data, (2)~applied AL in either setting (i.e., simulation or~practical), and (3) used AL in practical settings with human annotators. Questions are automatically skipped if participants are ineligible to answer based on prior responses. For example, participants who have never used AL will not be shown groups III--VI (see Figure~\ref{survey-structure-and-paths}c for different user groups). Thereby we aim to minimize the required effort for completing the survey, which is more appreciative of the participants' time, and also has been shown to positively affect the completion rates~\citep{liu:2018}. The full survey is provided in Appendix~\ref{sec:app:survey-questions}, including information on each question's format (i.e., free-form, multiple or single choice).

\subsection{Target Audience and Distribution}
The survey is targeted at the NLP domain and the minimum eligibility for participation is basic knowledge of supervised learning for NLP. 

We distributed the survey through various channels aiming to reach a broad audience: (1) mailing lists (ACL, ELRA corpora, ELRA SIGUL, Natural Language Processing DC, tada.cool); (2)~personalized email to a manually curated list of 601 individuals who co-authored papers on AL at major *CL venues between 2009 and 2024, and 9~additional personal contacts; (3)~common social media channels (LinkedIn, Bluesky, Twitter/X, Hugging Face Posts); (4)~outreach to annotation tool developers and providers, asking them to share the survey with their users. Supplementary details on distribution and implementation are provided in Appendix~\ref{sec:app:distribution-and-advertisements}.

The survey was open online to voluntary participants for 6 weeks, from December 15th, 2024, to January 26th, 2025. Calls for participation were shared in three waves: the initial invitation at the start of the survey period, a first reminder after two weeks, and a second reminder after four weeks.

\begin{figure}[t]
\centering
\includegraphics[width=0.487\textwidth]{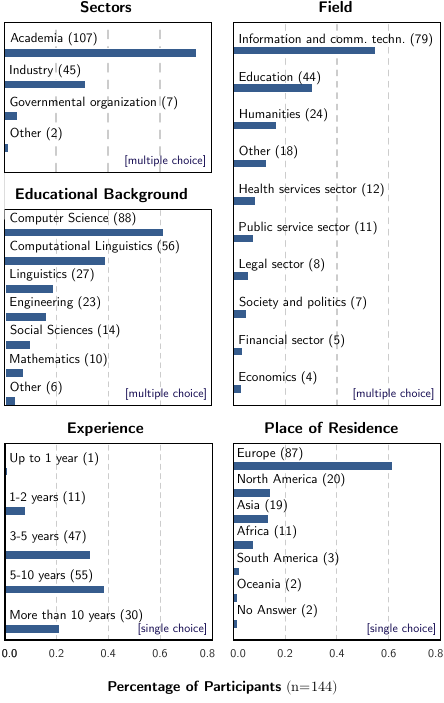}
\caption{Exploratory histograms of participants' work sectors, the field they are working in, educational background, work experience in machine learning/NLP, and primary place of residence (cf.~\protect\questionlink{VII.1}{VII.1-5}). The y-axis of each histogram represents the response options for the survey question, while the x-axis shows the proportion of participants who chose each response option.}
\label{background-summary}
\end{figure}

\subsection{Consent, Data Usage, and Privacy}

We implemented measures to ensure compliance with the EU General Data Protection Regulation. The survey was conducted anonymously, and participants consented to an anonymized release of their input. See Appendix \ref{app:data_privacy} for details.

%% file: part3-results.tex
\section{Results}

We first present the participant sample and then analyze the responses to RQ1--RQ4, deferring the discussion of RQ5 to Section \ref{sec:discussion}.

Table~\ref{tab:responses-full} in Appendix~\ref{app:additional-results} lists the response distributions for all 52 survey questions. In the remainder, we link specific arguments in the text to corresponding questions in Table~\ref{tab:responses-full} by question group and number (e.g., \question{I.1} for the first question).

\subsection{Participant Statistics}

Among 171 persons that navigated to the first question group, \surveyNumParticipants{}~completed the survey.\footnote{Compared to other surveys \cite{zhou:2022, subramonian:2023, michael:2023, blaschke:2024}, \surveyNumParticipants{}~full responses place us in the mid-range of participation.} This corresponds to a completion rate of 84\% of those that entered the content part of the survey. Incomplete responses were excluded from analysis.

Figure~\ref{background-summary} presents a demographic summary of the \surveyNumParticipants{}~participants. Most work in academia, but a substantial portion is employed in industry. Information and communications technology clearly leads the field of work, while the educational background is primarily in computer science and computational linguistics. Overall, the participants are very experienced in NLP, with the majority having between 3 to 10~years of experience. For recruiting participants, mailing lists were the most effective referral source~(51\%, cf.~\question{VII.7}).

A certain degree of selection bias is evident, and should be kept in mind when interpreting our results. Most notable is a geographical over-representation from Europe, despite considerable efforts to reach a diverse audience (see Appendix~\ref{sec:app:distribution-and-advertisements}; this limitation is also encountered by, e.g., \citet{hojer-etal-2025-research} despite a global recruitment strategy). Our targeted recruitment of authors may have introduced additional bias, but was deemed necessary to reach sufficient response rates on AL questions. Appendix \ref{app:conservative-selection} addresses potential overestimation.

\subsection{Data Annotation in Times of LLMs}
\label{sec:results-data-annotation-bottleneck}

First, we assess whether the community perceives a persistent need for data annotation (RQ1).
To capture perceived shifts in the relevance of annotated data, we ask those 138 respondents (out of 144) who had previously encountered difficulties due to a lack of annotated data (cf. \question{I.1}). We are interested in potential changes in difficulties given recent advancements in NLP.

\begin{figure}[t]
\centering
\includegraphics[width=0.48\textwidth]{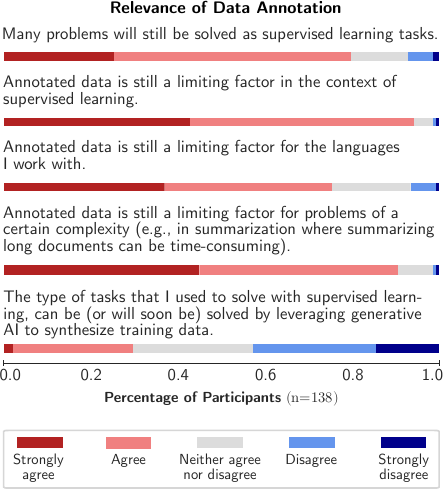}
\caption{Respondents' assessment of the relevance of data annotation in context of recent advancements in NLP using a 5-point Likert scale (cf.~\protect\questionlink{I.3}{I.3--7}).}
\label{relevance-of-data-annotation-likert}
\end{figure}

Figure~\ref{relevance-of-data-annotation-likert} illustrates the attitude on five questions about the relevance of data annotation currently. Overall, participants largely agree that many problems will still require supervised learning (80\%). Annotated data remains a limiting factor here; in general (94\%), for certain languages (75\%), and for problems of a certain complexity (91\%). Moreover, for their tasks, only few participants agree that generative AI for training data synthesis offers a viable solution (30\%).

\textit{Answer to RQ1: Supervised learning remains important in the age of LLMs. Data annotation is still a bottleneck, especially in demanding scenarios.}

\subsection{Overcoming Data Annotation Challenges}
\label{sec:overcoming-data-annotation-bottlenecks}

Next, we ask how AL is perceived for overcoming data annotation challenges compared to alternative methods (RQ2). 120 out of the 138 respondents who were eligible for RQ1 had used computational methods to answer this (cf.~\question{II.1}).

\begin{figure}[t]
\centering
\includegraphics[width=0.48\textwidth]{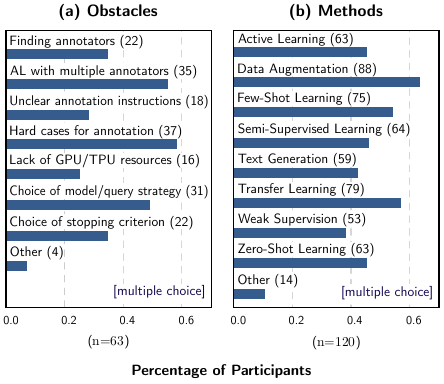}
\caption{Obstacles especially faced in the application of AL~(a) and participants' general experience with computational methods for a lack of annotated data (b). The y-axis represents the response options for the survey question, while the x-axis shows the proportion of participants who chose each response option.}
\label{fig:annotation-bottleneck}
\end{figure}

\paragraph{Adoption of active learning}
AL has been used by over half of the respondents (a total of 63; cf.~\question{II.2}), and the primary motivation was obtaining annotated data at minimal cost (87\%). Further goals included gaining practical experience (46\%), identifying difficult examples in annotation guidelines~(24\%), and improving data quality or using AL for research purposes (8\%; cf.~\question{III.2}).

When asking why participants had not chosen AL, 25\% had never heard of it, and 49\% cited insufficient expertise. There were accompanying methodological concerns, such as the expected implementation overhead (37\%), the lack of (or unawareness about) suitable annotation tools (32\%), the difficulty of estimating effectiveness upfront (12\%), and sampling bias (18\%). 9\% found it unsuitable due to specific project requirements and 14\% doubted AL’s effectiveness at all (cf.~\question{II.5}).

Despite these concerns, 54\% of participants who had not yet used AL consider its application in future projects and further 38\% indicated they are merely uncertain due to a lack of knowledge. Only one participant would explicitly not use AL (reasoning that it does not work well enough in practice; cf.~\question{II.8}), while four made its use dependent on the specific circumstances (16\%; cf.~\question{II.7}).

\paragraph{Alternative methods}
The current users of AL also report obstacles in data annotation with AL, illustrated in Figure~\ref{fig:annotation-bottleneck}a, which may lead them to opt for alternatives instead. The most pronounced challenge is handling hard annotation cases, followed by the involvement of multiple annotators. The third major hurdle is selecting an appropriate model and query strategy. In the other responses, the need for open, robust, and user-friendly AL tools was emphasized (cf.~\question{III.3}).

The use of computational alternatives to AL is widely dispersed across participants (see Figure~\ref{fig:annotation-bottleneck}b), but we observe that only data augmentation, transfer learning, and few-shot learning have been used more frequently than AL in our sample. In general, the alternative methods were considered successful in overcoming a lack of annotated data in slightly over half of the cases~(57\%, cf.~\question{II.4}). In comparison, recent AL-based annotation projects were considered successful in 91\% and effective in 67\% of cases~(see Section~\ref{sec:contemporary-active-learning} for more details).

\textit{Answer to RQ2: The strong openness to and perceived success of AL suggest its relevance for data annotation despite available alternatives.}

\subsection{Contemporary Active Learning}
\label{sec:contemporary-active-learning}

Moreover, we ask what contemporary setups of applied AL look like (RQ3). This section evaluates 33 AL projects\footnote{There is a near-equal split between academic (16) and industry/government respondents (20, including 3 overlaps).} with human annotators that were conducted within the last five years (from 2020 onwards; cf.~\question{III.5},~\question{IV.1}).

\begin{figure}[t]
\includegraphics[width=0.485\textwidth]{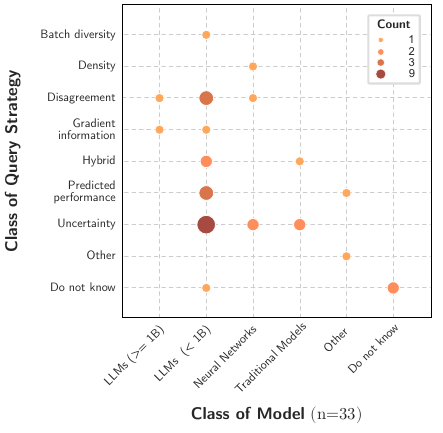}
\caption{Distribution of the reported combinations of model and query strategy.}
\label{fig:model-vs-query-strategy}
\end{figure}

\begin{figure}[t]
\includegraphics[width=0.485\textwidth]{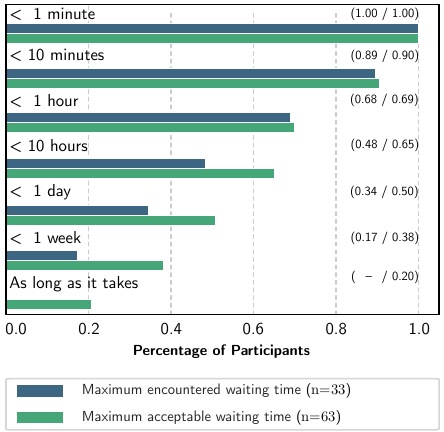}
\caption{Waiting times between two annotation cycles encountered in practical projects (cf.~\protect\questionlink{IV.4}{IV.4}) versus the maximum time respondents would accept (cf.~\protect\questionlink{III.4}{III.4}).}
\label{fig:waiting-times}
\end{figure}

\paragraph{Models, query strategies, and stopping criteria}

Figure \ref{fig:model-vs-query-strategy} provides an overview of models and query strategies used. The majority of projects relied on LLMs, with a clear preference for smaller models such as BERT. Notably, only two projects employed larger language models (in the context of our survey with 1B parameters and above). Uncertainty sampling emerged as the dominant query strategy, despite the wide range of alternatives.\footnote{\citet{zhang-etal-2022-survey} give an overview of query strategies.} Among the other options, we observe the tryout of multiple models and query strategies, as well as AutoML libraries for model selection (cf.~\questionlink{IV.2}{IV.2,3}). For stopping the AL process, most projects halted upon budget depletion (39\%), after evaluation of the model on a held-out gold standard (30\%) or via human assessment (3\%). Stopping at a fixed number of iterations or instances is less common~(12\%). Algorithmic stopping criteria are still rare, used in only 12\% of cases (cf.~\question{IV.5}).

\begin{figure}[t]
\includegraphics[width=0.485\textwidth]{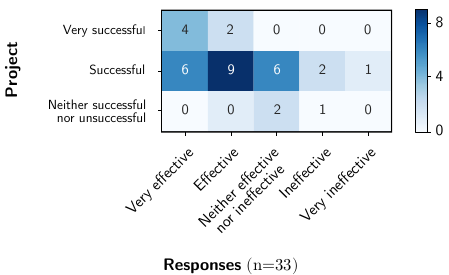}
\caption{The reported project success broken down by effectiveness of AL. 
Unselected options for project success are omitted (unsuccessful, very unsuccessful).}
\label{fig:project-success-vs-effectiveness}
\end{figure}

We also inquired regarding encountered waiting times and the maximum acceptable waiting times between annotation cycles, as illustrated in Figure~\ref{fig:waiting-times}. We observe a broad range reported across projects, from under one minute to up to one week. Contrary to our expectations, participants are willing to wait considerable times up to one week, over 50\% even as long as one day, suggesting that significant sampling times are neither unexpected nor problematic for them.

\paragraph{Annotation tools}

Given the availability of annotation tools with AL support, we asked participants about their usage. Interestingly, just over half of the projects used such a tool (17). These include \texttt{Argilla}~(4), \texttt{LabelSleuth} (3), \texttt{Prodigy} (2), \texttt{ActiveAnno} (1), \texttt{ALAMBIC} (1), \texttt{ALANNO} (1), \texttt{Label Studio} (1), and self-built solutions (4) (cf.~\questionlink{V.6}{V.6,7}).

\paragraph{Satisfaction with project}

Figure \ref{fig:project-success-vs-effectiveness} illustrates the perceived success of the annotation projects together with the perceived effectiveness of using AL. Overall, most projects were considered successful with an effective AL contribution. At the same time, especially for successful projects, it can be observed that the perception of AL effectiveness ranges from very high to very low (cf.~\question{V.11}, \question{V.12}).

Among those that reported a reduced effectiveness, the main reasons were the lack of suitable annotation tools (24\%; notably, 4 had used a tool with AL support) and the overhead of setup (21\%), which connects to insufficient knowledge in retrospect (17\%). Performance mismatches contributed, such as poor overall performance (21\%), a lack of effort reduction (9\%), inaccurate efficiency estimation (9\%), and mismatching project-specific requirements (6\%). In addition, respondents noted sampling bias (15\%) and dataset-model dependency (15\%). The other options stress concerns about setup time and effectiveness (cf.~\question{V.13}). Considering potential differences in terms of practical utility ``in the real world'' and for NLP researchers, we find that academic and industry/governmental respondents align closely on reasons for reduced effectiveness. Appendix~\ref{app:group-analysis} provides a detailed breakdown across sectors for the entire Section \ref{sec:contemporary-active-learning}.

Overall, 28 out of 33 AL practitioners stated they would use AL again in future projects. Two respondents restricted its use to specific scenarios, while another one remained uncertain. Only two respondents would not use AL again, noting that AL does not work well enough in practice (cf. \questionlink{V.14}{V.14,15}).

\textit{Answer to RQ3: The reported projects favor LLMs with uncertainty sampling. While project satisfaction is high in general, setup complexity (overhead, efficiency estimates, knowledge), general project risks (performance, effort reduction, sampling bias, model dependency, requirements), and unsuitable tools challenge AL adoption.}

\subsection{Anticipated Developments}
\label{sec:trends-and-next-steps}

We now turn to developments in AL that respondents anticipate (RQ4), starting with a reflection on four important advancements for NLP from recent years (see Figure~\ref{likert-trends-in-active-learning} in Appendix~\ref{app:additional-results}). We find that participants particularly agree on the impact of text embeddings (81\%), GPU-accelerated computing (67\%), and neural networks (59\%) on AL implementations. In contrast, the use of parameter-heavy LLMs was more controversial (36\% agreement).

We furthermore asked participants, through an open-ended question, about future AL trends in NLP they expect to shape the field. Table~\ref{table-next-developments} shows the results of our qualitative analysis. The majority of responses anticipated the integration of LLMs at various stages of AL. Further emphasis was put on improved instance selection. Responses also hint at expected advancements in tooling or improvements for low-resource languages and subjective tasks.

\textit{Answer to RQ4: Practitioners anticipate that the future of AL will revolve around increased and pragmatic LLM integration. Simultaneously, they call for fundamental improvements in existing query strategies and more accessible tooling.}

%% file: part4-related-work.tex
\section{Related Work}
\label{sec:related-work}

\paragraph{Progress in active learning research}

AL for NLP has substantially grown since its beginnings. Advancements have been reviewed in several literature surveys over time, focusing either on the overall picture \cite{olsson2009survey, zhang-etal-2022-survey}, or on certain sub-aspects such as deep neural networks and text classification \cite{schroeder2020surveyactivelearningtext}, or entity recognition \cite{kohl2024er}. The most recent academic developments are captured by \citet{xia-etal-2025-selection}, who surveyed LLM-based AL.

\input{table-next-developments}

\paragraph{Active learning in the era of LLMs}

The integration of LLMs and AL has been associated with a number of benefits. Among these are the outstanding performance of LLMs that can be boosted through AL even in demanding real-world data scenarios~\cite{ein-dor:2020}, resulting in a significant reduction of human annotation effort~\cite{shelmanov:2021, zhao:2020}\footnote{Few studies directly compare the effort reduction in AL between traditional models and LLMs, but \citet{romberg2022automated} show, e.g., that LLMs can halve the manual effort.}, the additional source of information that pre-training provides in the cold-start of AL~\cite{yuan:2020}, and capabilities such as few-shot learning~\cite{margatina-etal-2023-active,bayer:2024} and prompt-based strategies~\cite{li:2024}. LLMs have thus become the standard backbone models across tasks such as text classification~\citep{ein-dor:2020,yuan:2020,margatina:2021,galimzianova:2024}, sequence tagging ~\citep{shelmanov:2019,shelmanov:2021,luo:2023}, text summarization~\citep{tsvigun:2022b,li:2024,xia:2024} and machine translation~\citep{zeng:2019,zhao:2020,mendonca:2023,yuksel:2023}. Recent studies even suggest to entirely replace the human annotator with LLMs as cost-effective alternatives~\cite{xiao:2023, zhang:2023, kholodna:2024, liang:2024}. The reliability of LLM annotators, however, is still under debate. For example, \citet{baumann:2025} showed that with LLM annotators, incorrect statistical conclusions occurred in over a third of cases.

\paragraph{Existing literature on our research questions}

As anticipated in our motivation for conducting this community survey, the existing literature offers only limited answers to the research questions we address. Related work hardly questions the continued need for data annotation (RQ1) and the relevance of AL as a data-efficient method (RQ2). Our paper contributes by providing evidence to support this common assumption. Regarding contemporary AL setups in real-world use (RQ3), reports from actual AL applications or realistic simulations are rare \cite{tonneau:2022}. That said, a few studies do address challenges that we identified as persisting in practice, such as model-dataset dependency \cite{shelmanov:2021, tsvigun-etal-2022-towards, yuksel:2023, jelenic-etal-2023-dataset} and unstable performance and distribution mismatch \cite{li:2024}. The literature provides more extensive coverage of trends (RQ4), complementing the list of potential next steps in applied AL. There is initial research on LLMs as annotators \cite{zhang:2023, xiao:2023, liang:2024, kholodna:2024}, AL for few-shot learning or data synthesis \cite{margatina-etal-2023-active, bayer:2024, zhang:2023}, improving AL on low-resource languages through LLMs \cite{kholodna:2024}, cost-efficient LLMs for AL \cite{galimzianova:2024}, LLM-determined curricula for stable AL performance \cite{li:2024}, using AL to alleviate LLM hallucinations \cite{xia:2024}, handling human label variation \cite{wang-plank-2023-actor, van-der-meer-etal-2024-annotator} and annotation errors~\cite{weber:2023} with AL.

\paragraph{Prior community surveys}

Academic literature provides insights most of all from simulated experiments, including in literature surveys. Evaluations of practical applications are rare, and more importantly, can only represent the small portion of practical applications that have been documented in a scientific setting. Simulations, in contrast, provide only limited insights into the practical realities \cite{margatina-aletras-2023-limitations}. As a consequence, important aspects, such as negative results and practical obstacles, are likely missing. For the field of AL, there has only one community survey been conducted as early as 2009 by \citeauthor{tomanek-olsson-2009-web}. Given the significant progress in AL outlined above, our study provides timely new insights into practical implementations, going well beyond the earlier survey by introducing new question groups on the relevance of data annotation and AL, as well as on trends in AL for NLP.

%% file: table-next-developments.tex
\begin{table}[t]
\centering
\small
\begin{tabular}[!t]{@{}lr@{}}
\toprule
\textbf{\scshape Leveraging Language Models}\\
\midrule
\begin{minipage}[t]{0.47\textwidth}
\begin{enumerate}[label={\textbf{\arabic*.}}, itemindent=0pt, leftmargin=*, nosep,after=\vspace{+0.5\baselineskip},before=\vspace{-0.5\baselineskip}]
\setlength{\itemsep}{0px}
\item 
(Partially) using LLMs as annotators (8)
\item 
Increased use of LLMs in AL components (7)
\item 
Data synthesis during or before AL (4)
\item
Incorporate language model-based agents (3)
\item
Improved usage of embeddings (2)
\item
Using small efficient models (2)
\end{enumerate}
\end{minipage}\\
\midrule
\textbf{\scshape  Instance Selection}\\
\midrule
\begin{minipage}[t]{0.47\textwidth}
\begin{enumerate}[label={\textbf{\arabic*.}}, itemindent=0pt, leftmargin=*, nosep,after=\vspace{+0.5\baselineskip},before=\vspace{-0.5\baselineskip}]
\setlength{\itemsep}{0px}
\item 
More sophisticated query strategies (4)
\item 
Effectiveness over random sampling (2)
\item
More reliable, well-calibrated uncertainty scores (2)
\end{enumerate}
\end{minipage}\\
\midrule
\textbf{\scshape  Tooling}\\
\midrule
\begin{minipage}[t]{0.47\textwidth}
\begin{enumerate}[label={\textbf{\arabic*.}}, itemindent=0pt, leftmargin=*, nosep,after=\vspace{+0.5\baselineskip},before=\vspace{-0.5\baselineskip}]
\setlength{\itemsep}{0px}
\item 
Improvements in convenience (4)
\item 
Easier bootstrapping of an AL setup (2)
\end{enumerate}
\end{minipage}\\
\midrule
\textbf{\scshape Other}\\
\midrule
\begin{minipage}[t]{0.47\textwidth}
\begin{enumerate}[label={\textbf{\arabic*.}}, itemindent=0pt, leftmargin=*, nosep,after=\vspace{+0.5\baselineskip},before=\vspace{-0.5\baselineskip}]
\setlength{\itemsep}{0px}
\item 
AL for low-resource languages (2)
\item 
AL for subjective tasks (1)
\end{enumerate}
\end{minipage}\\
\bottomrule
\end{tabular}
\caption{Participant-identified trends in AL for NLP.
}
\label{table-next-developments}
\end{table}

%% file: part5-discussion.tex
\section{Discussion}
\label{sec:discussion}

We now first compare current perceptions to those from 2009 (RQ5)\footnote{The available data collection methodology does not allow us to claim representativeness of the NLP community for the two distinct samples. We therefore compare central findings rather than conducting longitudinal statistical analyses.}, then discuss the broader implications of our findings for AL as a field.

\paragraph{Comparison to \citet{tomanek-olsson-2009-web}}

The setup of AL has naturally shifted from maximum entropy approaches and support vector machines towards LLMs as the model of choice. Our survey indicates the popularity of smaller LLMs (in this study defined as models with less than 1B parameters). We hypothesize that the close or even on par performance these models can achieve with fine-tuning on many traditional supervised tasks (specifically encoders), combined with their lower hardware requirements, render larger models (i.e., LLMs with 1B parameters or more) computationally inefficient for practical application in AL. Larger LLMs may become the preferred choice in cases where a potentially minor performance increase is mission critical, or when no suitable smaller model exists (e.g., for languages where no strong encoder models are available). At the same time, uncertainty sampling remains the preferred query strategy. Although surprising at first glance, given the growing range of more sophisticated alternatives, this is in accordance with the literature on strong performance of uncertainty sampling \cite{shen-etal-2017-deep, margatina-etal-2022-importance, schroder-etal-2022-revisiting}. In terms of stopping criteria, practical constraints such as time and budget still largely outweigh information-theoretical grounds.

Revisiting the central challenges identified by \citet{tomanek-olsson-2009-web} (cf. Section~\ref{sec:introduction}), our current survey reveals that many of these issues remain relevant. Skepticism about the practical effectiveness of AL and the perceived overhead of implementing annotation interfaces continue to prevent community members from adopting AL. Similarly, users of AL continue to report difficulties related to sampling complexity and interface design. However, we did not observe notable challenges related to annotator waiting times, unlike in 2009; on the contrary, participants indicated a willingness to tolerate considerable delays. The shift to LLMs may have altered requirements here.

\textit{Answer to RQ5: LLMs have substantially transformed AL by serving as powerful backbone models. Nonetheless, key challenges to AL adoption persist despite considerable growth in AL research.}

\paragraph{Bridging the gap between research achievements and practical needs}

While LLMs offer options beyond serving as backbone models, these have not yet been integrated into the AL workflows reported by our participants. Approaches like prompt-based strategies \cite{li:2024} and in-context learning \cite{margatina-etal-2023-active} could make AL more accessible for less-expert users, and using AL for few-shot example selection \cite{bayer:2024} may further reduce annotation effort. Trends such as LLMs-as-annotators, however, require stronger theoretical foundations first before they can provide reliable support in practice~\cite{baumann:2025, tan-etal-2024-large}.

Despite progress on single issues, such as model-dataset dependency \cite{tsvigun-etal-2022-towards, jelenic-etal-2023-dataset} and performance stability \cite{li:2024}, research overall did not manage to sufficiently address the three overarching challenges to practical adoption that the community has raised:
\begin{enumerate}[leftmargin=0.5cm]
\setlength{\itemsep}{0pt}
    \item[(1)] {\em Excessive complexity of the AL setup}: Respondents who never had applied AL most often cite a lack of methodological expertise and implementation overhead as obstructing factors. AL users share the concerns regarding methodological choices. The ongoing introduction of new algorithms (e.g., query strategies) and, thus, possible setup combinations continually increases complexity.
    \item[(2)] {\em Risk of starting an AL project}: Respondents noted that the cost reduction through AL cannot be reliably estimated upfront. Its dependence on numerous variables makes a priori chosen setups prone to unsatisfactory outcomes \cite{lowell-etal-2019-practical}, possibly worse than random sampling~\cite{ghose-nguyen-2024-fragility}.
    \item[(3)] {\em Perceived insufficiency of tooling}: Despite considerable activity in the development of AL annotation tools, their progress is still perceived as insufficient. We suspect that this is partly due to the large problem space, making it difficult for tools to serve all tasks.
\end{enumerate}

As a starting point for addressing these challenges, we propose to:
\begin{enumerate}[leftmargin=0.5cm]
\setlength{\itemsep}{0pt}
\item[(1)] \textit{prioritize the reduction of existing complexity instead of devising new algorithms}, e.g., by refuting the effectiveness of established strategies;
\item[(2)] \textit{develop and continually research guidelines or heuristics for choosing AL components based on the setup's variables}, e.g., with a decision tree that specifies based on task and number of classes. More insights from practical use cases would also be immensely helpful to provide valuable points of comparison; and
\item[(3)] \textit{enhance the visibility of existing tools, and provide user feedback} that can be used by tool developers to refine their interfaces to ease the application of AL \cite{shnarch-etal-2022-label,jukic:2023}, especially for non-expert users~\cite{ras:2022}.
\end{enumerate}

%% file: part6-conclusions.tex
\section{Conclusions}

To gain comprehensive insights into AL in the era of LLMs, we conducted an online survey in the NLP community. Our findings suggest that AL is mostly viewed as a viable solution to the persistent challenge of data annotation. While recent research advancements have only partly permeated to AL applications yet, and the use of LLMs in AL is still on the rise, respondents are overall positive on the potential of LLMs in AL. We conclude with recommendations to improve three longstanding problems that hinder broader and more satisfying adoption, thereby hoping to strengthen AL further.

%% file: part-limitations.tex
\section*{Limitations}

\paragraph{Representativeness of the community survey}

Our results are descriptive for a sub-sample of the NLP community only, and therefore we cannot draw conclusions for the entirety of the community. As shown in Figure \ref{background-summary}, our distribution strategy led to a potential over-representation of academia as a work sector and European countries (primarily Germany). While the geographic reach is broad (covering 44 countries in total), a stronger representation from North America and Asia was expected.\footnote{\href{http://stats.aclrollingreview.org/submissions/geo-diversity/}{http://stats.aclrollingreview.org/submissions/geo-diversity/}, \href{https://www.marekrei.com/blog/geographic-diversity-of-nlp-conferences/}{https://www.marekrei.com/blog/geographic-diversity-of-nlp-conferences/}} Moreover, as our survey is distributed in English, this may introduce a language bias, possibly excluding respondents who are not proficient in English or decide against responding due to being uncomfortable responding in English. However, distributing the survey in multiple languages seems infeasible given the vast number of written languages.

\paragraph{Temporal bias in perception}

It is difficult to determine when participants formed their opinions on AL, as assessments are commonly based on past experiences that may not align with current advancements. Similarly, we cannot ensure that recent improvements in AL were acknowledged or influenced their opinion. As a result, the collected responses may be subject to a temporal bias, potentially pointing to issues that have since been improved upon or even resolved. However, our analysis of recent projects confirms that the three key challenges (setup complexity, cost reduction estimation, and tooling) persist, to a certain extent.

\paragraph{Constraints of active learning in low-resource language scenarios}

Naturally, AL can be suitable in scenarios in which \textit{annotated resources are scarce}, and yet \textit{non-annotated data is available}. However, limitations apply where using machine learning-based methods to tackle NLP tasks may be prevented by the almost complete lack of language-specific data (be it annotated or non-annotated) as well as non-existing NLP pipelines and tools. As some of our survey participants pointed out in the free-form text fields, non-machine learning techniques often have to be applied for those languages, and in consequence, AL cannot provide a solution here. We recognize that especially in the low-resource language field non-machine learning based approaches are still relevant today.

%% file: part-ethics.tex
\section*{Ethical Considerations}

To comply with data protection laws, we took the necessary steps to ensure compliance with the EU General Data Protection Regulation. Respondents granted their consent to the storage and analysis of their provided data, and the distribution of an anonymized version thereof, before starting the survey. To allow the NLP community to conduct extended analyses on the collected information and to uphold scientific transparency, we will publish the resulting dataset in a form that does not allow to infer conclusions about individuals.

Given the growing skepticism about data usage, particularly in the context of generative AI, and the voluntary nature of survey participation, we opted for a non-commercial license (i.e., CC BY-NC-SA 4.0\footnote{\href{https://creativecommons.org/licenses/by-nc-sa/4.0/}{https://creativecommons.org/licenses/by-nc-sa/4.0/}}). This approach ensures that the data can be accessed and used for research purposes while safeguarding respondents' privacy and addressing concerns about potential misuse.

%% file: part-acknowledgments.tex
\section*{Acknowledgments}

We thank all the great anonymous participants of the survey for their time and invaluable answers, which enabled this research. Likewise, we are grateful for everyone who supported us by distributing the survey, in particular Rabiraj Bandyopadhyay, David Berenstein, Ariel Gera, Alon Halfon, Philip H., Erik Körner, Gebriella Lapesa, Maximilian Maurer, Ines Montani, Paul Noirel, Julian Oestreich, Myrthe Reuver, Arij Riabi, Stefan Schweter, Artem Shelmanov, Eyal Shnarch, Manuel Tonneau and Akim Tsvigun. We are equally grateful to all anonymous supporters. Thanks to all tool authors and friends from the AL community for their support. We sincerely appreciate the contributions of all who helped, even if not mentioned by name.

We thank Ranjit Singh of \textit{GESIS Survey Methods Consulting} for his expertise and feedback on early drafts of the survey, Harry Scells and Erik Körner who reviewed and tested the survey, Gerhard Heyer and Ingolf Römer for supporting the GDPR agreements, and \href{https://bildungsportal.sachsen.de/}{Bildungsportal Sachsen} for providing the LimeSurvey infrastructure, Constantin Hammer, Luisa Köhler, and Jens Syckor for supporting us with the GDPR-compliance in their roles as data protection officers, and Ana Lisboa Cotovio for support in the data collection for Figure~\ref{conference-publications-versus-active-learning}.

Parts of this paper have been financially supported by the German Federal Ministry of Research, Technology and Space (BMFTR) and by Sächsische Staatsministerium für Wissenschaft, Kultur und Tourismus in the programme Center of Excellence for AI-research ``\href{https://scads.ai/}{Center for Scalable Data Analytics and Artificial Intelligence Dresden/Leipzig}'', project identification number: ScaDS.AI. Further parts of this work have been conducted within the \href{https://coral-nlp.github.io}{CORAL project} which was funded by the German Federal Ministry of Research, Technology, and Space (BMFTR) under the grant number 16IS24077A. Responsibility for the content of this publication lies with the authors.

%% file: part-appendix.tex
\appendix

\section*{Appendix}

\section{Active Learning Publications}
\label{sec:app:publication-identification}

To obtain the data for Figure \ref{conference-publications-versus-active-learning}, we start from the full ACL anthology in BibTeX format, including abstracts (January 14th, 2026). In a first step, only publications published in the context of AACL, ACL, COLING, EACL, EMNLP, NAACL, CL, and TACL between 2009 and 2025 were kept. This includes affiliated workshops. The resulting number of papers constitutes the \textit{total number of publications}. Subsequently, we filtered publications that contain ``active'' in title or abstract. If ``active'' is directly followed by ``learning'', the publication is kept without further checks, in case of other next words, we manually checked the publication's relevance.\footnote{This way, we account for wording variations such as ``active fine-tuning'' or ``active sampling.''} The resulting number of papers constitutes the \textit{total number of AL publications}.

The numbers for 2025 must be interpreted with caution, as the ACL Anthology had not yet been fully updated with all entries at the time of analysis (e.g., the proceedings of AACL, some workshops).

\section{Survey Implementation}
\label{sec:app:distribution-and-advertisements}

\subsection{Survey Distribution and Advertisement}
We chose a diverse distribution strategy, covering mailing lists, direct contacts, social media, and service providers. While some overlap between these channels may exist, we utilize all of them to ensure a wide outreach.

\paragraph{Mailing lists}
The \textbf{ACL mailing list} covers all current members of the Association of Computational Linguistics -- about 10,400 as of end of 2024. \textit{The worldwide recipients are researchers from academia and industry likewise.}\footnote{The call to the ACL mailing list was sent out only once, due to two reasons: first, a manual inspection period of about two weeks by the administrators before distribution limited us to requesting only two mailings during the survey period; second, our second request was deleted for unclear reasons.} The \textbf{ELRA corpora-list} is managed by the European Language Resources Association. It serves as a platform for sharing information related to linguistic resources, therefore \textit{providing a way to reach researchers and practitioners involved in data annotation in their daily work}. The exact number of recipients on the mailing list is not public; however, it is widely assumed as reaching a substantial audience. The members of \textbf{ELRA SIGUL-list} are professionals involved in the development of language resources and technologies for under-resourced languages. By utilizing this mailing list, we aim to \textit{gain a more diverse perspective on supervised learning and AL for lesser-studied languages}, trying to mitigate potential biases in data collection driven by the uneven availability of language resources. The exact number of recipients on the mailing list is not public. \textbf{Natural Language Processing Data Community} is a google group\footnote{\href{https://groups.google.com/a/datacommunitydc.org/g/nlp}{https://groups.google.com/a/datacommunitydc.org/g/nlp}} for \textit{anyone interested in NLP including computational linguists, data scientists, and software engineers}. It counts 341 members as of the end of 2024. The \textbf{tada.cool initiative}\footnote{\href{https://sites.google.com/view/polsci-ml-initiative/talks}{https://sites.google.com/view/polsci-ml-initiative/talks}} is a community of \textit{interdisciplinary researchers, with a focus on machine learning and NLP for social science applications}. We reached 748 members via the slack channel.

\paragraph{Social media}
To gather more applied perspectives, we identified several chat communities on NLP and machine learning topics (e.g., PyTorch Lightning, Hugging Face Discord, and Hugging Face Posts) and sought approval to share the survey on these platforms. Approval was granted for the latter two, and the call for participation was subsequently posted. We also posted participation requests on several social media platforms (LinkedIn, X, and Bluesky), which are frequented by both audiences. These posts were re-shared by non-author accounts, further extending the survey's reach.

\paragraph{Researchers working on AL in NLP}
We curated a list of researchers who co-authored one or more publications on AL in major NLP venues over the past 15 years. To compile this list of experts, we started from the AL publications identified as described in Appendix \ref{sec:app:publication-identification}. We manually extracted the email addresses from publications that mentioned AL directly in the title. The approach also includes papers from workshops to capture more practical application experiences. We contacted all 601 identified individuals by a personalized mail. In 139 cases, the extracted e-mail addresses were no longer functional. We added 9 further personal contacts to complement the list.

\paragraph{Annotation tool providers}
As an additional strategy to incorporate practical insights from individuals involved in dataset creation efforts---potentially supported by AL---we reached out to individuals and companies responsible for annotation tools with AL solutions for text annotation.

To identify relevant tools, we supplemented a recent survey of annotation tools \citep{borisova:2024} with additional tools we encountered through years of research in AL. We contacted: ActiveAnno \cite{wiechmann-etal-2021-activeanno}, AL Toolbox \cite{tsvigun:2022}, Argilla\footnote{\href{https://github.com/argilla-io/argilla}{https://github.com/argilla-io/argilla}}, AWS Sagemaker Ground Truth and Comprehend, BioQRator \cite{Kwon2013BioQRatorA}, CleanLab\footnote{\href{https://github.com/cleanlab/cleanlab}{https://github.com/cleanlab/cleanlab}}, INCEpTION \cite{klie-etal-2018-inception}, Labelbox\footnote{\href{https://labelbox.com/}{https://labelbox.com/}}, Label Sleuth \cite{shnarch:2022}, Label Studio\footnote{\href{https://labelstud.io/}{https://labelstud.io/}}, MITRE Annotation Toolkit (MAT)\footnote{\href{https://sourceforge.net/projects/mat-annotation/}{https://sourceforge.net/projects/mat-annotation/}}, and POTATO \cite{pei-etal-2022-potato}. We also contacted several companies and services via their official communication channels, asking them to forward the survey to their users. As an incentive, potential insights that could be gained into the actual needs of users in supervised learning and AL were emphasized. Although this approach yielded limited feedback, it nevertheless contributed to the overall outreach. Positive responses were received from AL Toolbox, which forwarded the call to some of their annotators; Argilla, which permitted the call to be shared in the Hugging Face Discord channel; Explosion, which shared the call on LinkedIn; Labelbox, which forwarded the call internally; and Label Sleuth, which forwarded the call to their users.

In retrospect, we believe the limited engagement from tool providers and companies may be attributed to a combination of factors: an insufficiently compelling appeal or reward for sharing that we could provide from our side, the possibility that the survey was not a suitable fit for their interests, and the timing of the survey launch.

\subsection{Survey Portal \& Execution}

The survey was conducted using \href{https://www.limesurvey.org}{LimeSurvey} in version 3.27.4. Participants could revise their answers at any time; they could navigate back and forth, as long as the survey had not been completed.

\subsection{Participation and Completion Rates}
The survey remained open for six weeks. According to question~\question{VII.7}, mailing lists were the most effective referral source, with noticeable spikes in activity following each call (with declining impact). Given this pattern, and after three waves of invitations, we did not expect a significant increase in responses by keeping the survey open longer.

Given that we have no information about the number of subscribers to the various mailing lists, or who we reached via social media or other means, it is impossible to determine how many individuals received the invitation to participate in the survey. According to LimeSurvey statistics, 1124 individuals clicked the survey link. Of these, 171 began the main content of the survey, and 144 completed it.

\subsection{Data Protection}
\label{app:data_privacy}

In consultation with data protection officers, we implemented measures to ensure compliance with the EU General Data Protection Regulation. Following it's principle of data minimization, we collected only data necessary for our research objectives.~The survey was conducted anonymously, with the option to provide an email address to receive the survey evaluation afterward. Free-form text fields are a gray area, as participants may potentially enter personally identifiable information, particularly when combined with demographic data collected.~To meet legal obligations, we informed participants about how any personal data would be processed, and concluded a joint-controller agreement among the institutions involved.

\subsection{Anonymization for Dataset Release}

For public release, text answers were manually processed as follows. We started from the raw dataset, excluding the separately stored email addresses. For each field, we replaced all mentions that could reveal the respondents' identities with placeholders, including URLs, links, email addresses, single words, sentences, and paragraphs. We also applied this to language mentions, as they could contribute to de-anonymization when combined with other fields (e.g., in case of languages that few people in the NLP community are researching). In this context, we took a conservative approach to V.2 and V.7 by removing all answers to preserve anonymity. For question items containing highly specific information that may allow to draw conclusions about the respective participant, we replaced the entries with broader categories:
\textbf{IV.1 (year):} 2020--2024, 2015--2019, 2010--2014, 2005--2009;
\textbf{V.3 (hours):} up to 20, 21--50, 51--100, 101--500, more than 500;
\textbf{V.4 (instance count):} up to 100, 101--500, 501--1000, 1001--10,000, more than 10,000,~N/A;
\textbf{V.4 (instance type):} documents, sentences, tokens, other,~N/A;
\textbf{VII.5 (countries):} Africa, Asia, Europe, North America, Oceania, South America.

Responses and demographics are released separately and randomly shuffled to prevent re-merging.

\section{Full Survey}
\label{sec:app:survey-questions}

\subsection*{\small Data Annotation Bottleneck \& Active Learning for NLP in the Era of LLMs}

{\small
\noindent The success of Natural Language Processing (NLP) often depends on the availability of (high-quality) data. In particular, \textbf{the costly manual annotation of text data has posed a major challenge since the early days of NLP}. To overcome the data annotation bottleneck, a number of methods have been proposed. One prominent method in this context is \textbf{Active Learning}, which aims to minimize the set of data that needs to be annotated.\\

\noindent However, the development of Large Language Models (LLMs) has changed the field of NLP considerably. For this reason, it is of huge interest to us working in this field (both in research and in practical application) to understand \textbf{if and how a lack of annotated data is still affecting NLP today}.\\

\noindent At the center of this survey is Active Learning, which was last surveyed in a \href{https://aclanthology.org/W09-1906.pdf}{web survey in 2009}. Fifteen years later, we aim to reassess the current state of the method from the user's point of view. Besides inquiring where Active Learning is used, we also ask where it is not used in favor of other methods. Moreover, we want to understand which computational methods the community considers most useful to overcome a lack of annotated data.\\

\noindent \textbf{\textit{The survey is conducted solely for non-commercial, academic purposes. It specifically targets participants who are or have been involved in supervised machine learning for NLP. Knowledge about Active Learning is not required. Filling out the survey will take you approximately 15 minutes.}}\\

\noindent \textbf{Why should I invest my time in this survey?} We need your collective expertise in the field of NLP, Supervised Machine Learning, or Active Learning, to understand how recent advancements, such as LLMs, have changed the long-standing data annotation bottleneck. The results of this survey will help the community to better understand the state and open issues of contemporary Active Learning, and incorporate these insights into research and development of new methods and technologies. To this end, a study presenting and discussing the results of this survey will be published as an open access publication. If you wish to be notified upon publication, you can optionally enter your email address at the end of the survey.\\

\noindent \textbf{What is Active Learning?} Active Learning is a method to create a small but meaningful annotated dataset with the goal of minimizing the annotation effort. It is an iterative cyclic method between a human annotator and a learning algorithm. In each iteration, an instance selection strategy (also referred to as query or acquisition strategy) is used to select data points that are considered particularly useful to be annotated next. These can be, for example (among many other strategies), the instances for which a machine learning model is most uncertain.\\

\noindent The survey is initiated by the researchers Christopher Schröder (Institut für Angewandte Informatik e. V.), Julia Romberg (GESIS - Leibniz Institute for the Social Sciences), and Julius Gonsior (TU Dresden). If you have any questions, please contact us via activelearningsurvey2024@gmail.com.

}

\subsection*{\small Consent, Data Usage, and Privacy}

\small{
\textbf{Declaration of Consent}\\
\noindent\textit{[link to declaration of consent]}\\

\noindent We need your consent before we can start collecting data for the survey.
\begin{enumerate}
    \item[()] Yes, I would like to participate in this survey.
    \item[()] No, I would not like to participate in this survey.
\end{enumerate}

\noindent\textbf{Data Usage and Privacy Information:} We will store your provided answers on a server located in \textit{[anonymized for review]}. After the survey has been conducted, we will process the collected data in order to investigate the research questions presented above to investigate if and how a lack of annotated data still affects the field of NLP in 2024. Our goal is to make the collected data available to the community under a \href{https://creativecommons.org/licenses/by-nc-sa/4.0/}{non-commercial share-alike CC BY-NC-SA 4.0 license}. \textit{\textbf{The public dataset will be completely anonymous.}} Any information in free text responses that could identify specific respondents will be removed before publication of the dataset. The processing of any personal data is detailed in privacy information in compliance with GDPR. This includes information on the purpose of data collection, how long we retain your data, your rights regarding your data, and how to contact us with any concerns.\\

\noindent We need your confirm that you have read and consented to the data usage and privacy information.
\begin{enumerate}
    \item[()] I have read the information on data usage and privacy information and consent that my answers given in this survey will be stored and analyzed for research purposes as described above. I agree that my contribution will be made available to the community as part of a public dataset under CC BY-NC-SA 4.0 license. I hereby grant the authors permission to distribute my data under these terms. It will be ensured that no conclusions about individuals can be inferred from this data.
\end{enumerate}
}

\subsection*{\small I. Relevance of Annotated Data in Light of Recent Advancements in NLP}

{\small
This survey focuses on supervised learning in NLP, where annotated data is used to train machine learning models.

\begin{enumerate}
    \item {[sc,m]}\footnote{Question properties: single choice (sc), multiple choice (mc), mandatory (m), and display conditional on previous response (c).} When applying NLP methods, have you ever encountered difficulties due to a lack of annotated data (e.g., you had to build a new dataset, without which the desired goal was unattainable)?
    \begin{enumerate}
        \item Yes
        \item No \textit{[directs participants to part VII.]}
    \end{enumerate}
\end{enumerate}

\noindent You stated that you have had difficulties due to a lack of data.

\begin{enumerate}
    \item[2.] {[mc,m]} Under what circumstances did you encounter difficulties due to lack of annotated data? Select all that apply.
    \begin{enumerate}
        \item When working on following under-resourced languages: \textit{[text input]}
        \item When working on a certain task (e.g., summarization): \textit{[text input]}
        \item When working with a specific requirement (e.g., on a problem with many classes): \textit{[text input]}
        \item Other: \textit{[text input]}
    \end{enumerate}
\end{enumerate}

\noindent Recent advancements, such as LLMs and Generative AI, have profoundly changed the NLP landscape. Considering these developments, do you think that a lack of annotated data is \textbf{currently} a problem? Please indicate below, how much you agree or disagree with each of these statements.

\begin{enumerate}
    \item[3.] {[sc,m]} The type of tasks that I used to solve with supervised learning, can be (or will soon be) solved by leveraging generative AI to synthesize training data.
    \begin{itemize}
        \item Strongly agree, Agree, Neither agree nor disagree, Disagree, Strongly disagree \textit{(5-point Likert scale)}
    \end{itemize}
    \item[4.] {[sc,m]} Many problems will still be solved as supervised learning tasks.
    \begin{itemize}
        \item Strongly agree, Agree, Neither agree nor disagree, Disagree, Strongly disagree \textit{(5-point Likert scale)}
    \end{itemize}
    \item[5.] {[sc,m]} Annotated data is still a limiting factor in the context of supervised learning.
    \begin{itemize}
        \item Strongly agree, Agree, Neither agree nor disagree, Disagree, Strongly disagree \textit{(5-point Likert scale)}
    \end{itemize}
    \item[6.] {[sc,m]} Annotated data is still a limiting factor for the languages I work with.
    \begin{itemize}
        \item Strongly agree, Agree, Neither agree nor disagree, Disagree, Strongly disagree \textit{(5-point Likert scale)}
    \end{itemize}
    \item[7.] {[sc,m]} Annotated data is still a limiting factor for problems of a certain complexity (e.g., in summarization where summarizing long documents can be time-consuming).
    \begin{itemize} 
        \item Strongly agree, Agree, Neither agree nor disagree, Disagree, Strongly disagree \textit{(5-point Likert scale)}
    \end{itemize}
\end{enumerate}
}

\subsection*{\small II. Overcoming a Lack of Annotated Data}

{\small
Annotated data is required for supervised learning. In this section, we ask about methods to overcome a lack of data in such situations, and about your experiences with them.

\begin{enumerate}
    \item {[sc,m]} Have you ever used a computational method to overcome a lack of annotated data?
    \begin{enumerate}
        \item Yes
        \item No
    \end{enumerate}
    \item {[mc,m,c:II1]} Please choose all computational methods that you have ever used to overcome a lack of annotated data.
    \begin{enumerate}
        \item Active Learning (e.g., to efficiently annotate data)
        \item Data augmentation (e.g., to use existing training instances more efficiently)
        \item Few-shot Learning (e.g., to achieve competitive results with only a few instances)
        \item Semi-supervised Learning (e.g., use already annotated data for automatic labeling of training instances)
        \item Text generation (e.g., to generate training instances)
        \item Transfer learning (e.g., to use the pre-existing knowledge of already trained models)
        \item Weak supervision (e.g., to programmatically obtain (pseudo-)annotated instances using, e.g., labeling functions)
        \item Zero-shot (e.g., to work without any training data at all)
        \item Other: \textit{[text input]}
    \end{enumerate}
    \item {[mc,m,c:II1]} What are the reasons why you have never used a computational method to overcome a lack of annotated data? Please choose all options that apply to you.
    \begin{enumerate}
        \item I have never heard of these methods.
        \item I never coordinated an annotation project.
        \item We had enough human resources to annotate our data.
    \end{enumerate}
\end{enumerate}

\noindent You stated that you have used a method other than Active Learning.

\begin{enumerate}
    \item[4.] {[sc,m,c:II2]} Did you consider your chosen computational method(s) to overcome a lack of annotated data to be successful?
    \begin{itemize}
        \item Strongly agree, Agree, Neither agree nor disagree, Disagree, Strongly disagree \textit{(5-point Likert scale)}
    \end{itemize}
    \item[5.] {[mc,m,c:II2]} What are the specific reasons for not applying Active Learning? Please select all that apply.
    \begin{enumerate}
        \item Never heard of Active Learning.
        \item Insufficient expertise/knowledge. 
        \item Did not want to spend overhead in implementing an Active Learning-based annotation environment.
        \item Did not know of any suitable annotation tools that easily integrate Active Learning.
        \item Did not meet my project's specific requirements.
        \item Wanted to avoid sampling bias in the corpus.
        \item Was not convinced that Active Learning would reduce annotation cost.
        \item Did not use Active Learning because I couldn't estimate upfront its impact on reducing annotation costs.
        \item Other: \textit{[text input]}
    \end{enumerate}
    \item[6.] {[m,c:II5]} Why did Active Learning not fit your specific requirements? \textit{[text input]}
    \item[7.] {[sc,m,c:II2]} Would you consider applying Active Learning in future annotation projects?
    \begin{enumerate}
        \item Yes
        \item No
        \item I don’t know enough about Active Learning to answer this question.
        \item Only under specific circumstances: \textit{[text input]}
    \end{enumerate}
    \item[8.] {[mc,m,c:II7]} What are the reasons why you would not apply Active Learning in future annotation projects?
    \begin{enumerate}
        \item Active Learning has become obsolete nowadays.
        \item Active Learning is a useful concept in theory but does not work well enough in practice.
        \item Other: \textit{[text input]}
    \end{enumerate}
\end{enumerate}

\noindent\textit{[directs participants who never used Active Learning to part VII.]}
}

\subsection*{\small III. Active Learning -- General}

{\small
In this section, we want to learn more about your experience with Active Learning. 

\begin{enumerate}
    \item {[mc,m]} What is your expertise with Active Learning?
    \begin{enumerate}
        \item I have been part of an Active Learning workflow as an annotator.
        \item I have organized an Active Learning annotation workflow.
        \item I have knowledge about instance selection strategies in Active Learning but use pre-built solutions.
        \item I have implemented Active Learning strategies or workflows.
        \item I have researched Active Learning.
    \end{enumerate}
    \item {[mc,m]} What was your primary motivation to use Active Learning?
    \begin{enumerate}
        \item Obtain annotated data at minimal annotation costs 
        \item Gather experience with practical applications of Active Learning 
        \item  Test your annotation guidelines to identify difficult examples
        \item Other: \textit{[text input]}
    \end{enumerate}
\end{enumerate}

\noindent In the following questions, we sometimes ask about \textit{(annotation) instances}, by which we mean the entities to be annotated, such as documents, words, or sentences. 

\begin{enumerate}
    \item[3.] {[mc,m]} What obstacles have you encountered when applying Active Learning?
    \begin{enumerate}
        \item Finding annotators that achieve satisfactory performance
        \item Performing Active Learning with multiple annotators (e.g., distributing work, resolving disagreement)
        \item Unclear instructions how to annotate (lack of annotation guidelines)
        \item Hard cases for annotation (undecidable, ambiguous, subjective) 
        \item Lack of resources (GPU/TPU)
        \item Choice of model and instance selection algorithm
        \item Choice of stopping criterion
        \item Other: \textit{[text input]}
    \end{enumerate}
    \item[4.] {[sc,m]} Between two iterations of Active Learning, a new model is usually trained. This new model is then used to select new examples for annotation. What is the maximum time you would be willing to wait between two annotation cycles?
    \begin{enumerate}
        \item Less than a minute
        \item 1 minute to less than 10 minutes
        \item 10 minutes to less than 1 hour
        \item 1 hour to less than 10 hours
        \item 10 hours to less than 1 day
        \item 1 day to less than 1 week
        \item I will wait long as it takes
    \end{enumerate}
    \item[5.] {[mc,m]} In which scenario(s) have you worked with Active Learning?
    \begin{enumerate}
        \item In a lab setting with simulated annotators
        \item In an applied setting with human annotators
    \end{enumerate}
\end{enumerate}

\noindent\textit{[directs participants that never used Active Learning in an applied setting to part VI.]}
}

\subsection*{\small IV. Active Learning -- Methodological Setup}

{\small
The following questions relate to type and scope of a previous annotation project in which you employed Active Learning.\\

\noindent In case you performed more than one such annotation project, please think of your \textbf{most recent project} with Active Learning when answering the following questions.

\begin{enumerate}
    \item {[m]} In which year did the project start?  \textit{[text input]}
    \item {[sc,m]} Which type of machine learning model was trained during Active Learning?
    \begin{enumerate}
        \item Classic (Naive Bayes, Support Vector Machine, Conditional Random Fields,~...)
        \item Neural Network (Convolutional Neural Network, Long Short-term Memory, ...)
        \item BERT era Language Models such as BERT, GPT-2, RoBERTa, ...
        \item Large Language Models such as Phi-3, Llama, Mistral, Gemma, Gemini, GPT-3/4, ...
        \item Do not know
        \item Other: \textit{[text input]}
    \end{enumerate}
    \item {[sc,m]} What instance selection strategy\footnote{Tooltip: The most salient component of an Active Learning setup is the instance selection strategy, which decides on the instances to be labeled next.} have you used during Active Learning? If you have tried different strategies, please indicate the strategy that you eventually decided on.
    \begin{enumerate}
        \item Uncertainty-based\footnote{Tooltip: Selects instances for which the model is most uncertain. Examples are least-confidence, entropy and margin-sampling.}
        \item Disagreement-based\footnote{Tooltip: Uses multiple models to select instances based on disagreement between model outputs. A popular example is query-by-committee.}
        \item Gradient information-based\footnote{Tooltip: Selects instances based on their impact on the model. The impact is measured by, e.g., the norm of the gradients. A popular example is expected gradient length.}
        \item Performance prediction-based\footnote{Tooltip: Selects instances that have the most potential of reducing future errors. Examples are policy-learning strategies with reinforcement learning or imitation learning, and cartography Active Learning.} 
        \item Density-based\footnote{Tooltip: Selects instances that are representative of dense regions in the embedding space. Density-based representatives can be selected, e.g., based on n-gram counts.}
        \item Discriminative-based\footnote{Tooltip: Selects instances that differ from already annotated instances. Examples are selecting discriminative-based representatives based on rare words or a lesser similarity to the already annotated instances, as well as the use of discriminative Active Learning.}
        \item Batch diversity\footnote{Tooltip: Selects a batch of instances that are diverse. Examples are coresets, BADGE and ALPS.}
        \item Hybrid (Combines different of the aforementioned concepts.)
        \item Do not know
        \item Other: \textit{[text input]}
    \end{enumerate}
    \item {[sc,m]} Between two iterations of Active Learning, a new model is usually trained. This new model is then used to select new examples for annotation. How long was the resulting waiting time for the annotator in your scenario on average? Please select the matching time interval.
    \begin{enumerate}
        \item We did not retrain
        \item Less than a minute
        \item 1 minute to less than 10 minutes
        \item 10 minutes to less than 1 hour
        \item 1 hour to less than 10 hours
        \item 10 hours to less than 1 day
        \item 1 day to less than 1 week
        \item Do not know
    \end{enumerate}
    \item {[sc,m]} How did you decide when to stop the Active Learning process? 
    \begin{enumerate}
        \item Evaluation of the learned model on a held-out gold standard 
        \item Money and/or time available for annotation were depleted
        \item All relevant documents annotated 
        \item The stopping time was indicated by an algorithmic stopping criterion (for example based on the current model’s performance)
        \item Do not know
        \item Other: \textit{[text input]}
    \end{enumerate}
\end{enumerate}
}

\subsection*{\small V. Active Learning -- Task \& Annotation Setup}

{\small
The following questions relate to type and scope of a previous annotation project of your choice, where Active Learning was employed.\\

\noindent In case you performed more than one such annotation project, please think of your \textbf{most recent project} with Active Learning when answering the following questions.

\begin{enumerate}
    \item {[sc,m]} What specific NLP task did you collect annotations for?
    \begin{enumerate}
        \item Automatic speech recognition
        \item Coreference resolution
        \item Chunking
        \item Information extraction
        \item Language generation
        \item Language understanding
        \item Machine translation
        \item Morphological analysis
        \item Named entity recognition
        \item Part-of-speech tagging
        \item Question answering
        \item Relation extraction
        \item Semantic similarity
        \item Sentiment analysis
        \item Syntactic parsing
        \item Summarization
        \item Text categorization
        \item Word segmentation
        \item Word sense disambiguation
        \item Other: \textit{[text input]}
    \end{enumerate}
    \item {[mc,m]} What was the language of the texts that were annotated? (If multilingual, select multiple options.)
    \begin{enumerate}
        \item Arabic
        \item English
        \item French
        \item German 
        \item Hindi 
        \item Mandarin 
        \item Spanish 
        \item Other: \textit{[text input]}
    \end{enumerate}
    \item {[sc,m]} How much time was spent on annotation in total? In the case of multiple annotators, please enter the total sum of hours worked by the annotators.
    \begin{enumerate}
        \item \textit{[text input]} hours
        \item Do not know
    \end{enumerate}
    \item {[m]} What was the size of the resulting annotated corpus in terms of annotated instances? Please provide the annotation instance type in brackets, e.g. 1000 (tokens), 100 (sentences), or 10 (documents)..  \textit{[text input]}
    \item {[mc,m]} Who were your annotators?
    \begin{enumerate}
        \item Annotation was done by the project coordinators or other project members (data scientist / linguist roles or similar)
        \item Domain experts
        \item Non-domain experts
        \item Other: \textit{[text input]}
    \end{enumerate}
    \item {[sc,m]} Did you use an annotation tool with Active Learning support?
    \begin{enumerate}
        \item Yes
        \item No
    \end{enumerate}
    \item {[sc,m,c:V6]} Which annotation tool with Active Learning support did you use?
    \begin{enumerate}
        \item ActiveAnno
        \item Argilla
        \item AWS Sagemaker Ground Truth
        \item AWS Comprehend
        \item BioQRator
        \item INCEpTION
        \item Labelbox
        \item Label Studio
        \item MAT
        \item Potato
        \item Prodigy
        \item Other: \textit{[text input]}
    \end{enumerate}
    \item {[sc,m]} In annotation, what did you do with data points that were ambiguous (i.e., caused disagreement between multiple annotators)?
    \begin{enumerate}
        \item Deleted them
        \item Assigned one of two labels
        \item Not applicable, each instance had at most one annotation.
        \item Other: \textit{[text input]}
    \end{enumerate}
    \item {[sc,m]} While using Active Learning, did you have to change the annotation schema due to challenging examples surfaced during the annotation process (eg, add new classes in a classification task)?
    \begin{enumerate}
        \item No
        \item Yes
    \end{enumerate}
    \item {[sc,m,c:V9]} What did you do with previously annotated examples?
    \begin{enumerate}
        \item Previously annotated examples were re-annotated.
        \item Previously annotated examples were left unchanged. 
        \item Other: \textit{[text input]}
    \end{enumerate}
    \item {[sc,m]} Did you consider your annotation project to be successful?
    \begin{itemize}
        \item Very successful, Successful, Neither unsuccessful nor successful, Unsuccessful, Very unsuccessful \textit{(5-point Likert scale)}
    \end{itemize}
    \item {[sc,m]} Was the use of Active Learning as effective as intended?
    \begin{itemize}
        \item Very effective, Effective, Neither effective nor ineffective, Ineffective, Very Ineffective \textit{(5-point Likert scale)}
    \end{itemize}
    \item {[mc,m,c:V12]} What were the reasons that reduced the effectiveness of Active Learning in your case of application? Please choose all options that apply.
    \begin{enumerate}
        \item In retrospect, I had insufficient expertise/knowledge.
        \item Overhead in setting up an Active Learning-based annotation environment.
        \item Lack of suitable annotation tools that easily integrate Active Learning.
        \item In retrospect, Active Learning did not meet my project's specific requirements.
        \item Active Learning created a sampling bias in the corpus.
        \item Active Learning did not reduce annotation cost.
        \item Active Learning did not work well in my scenario.
        \item My upfront estimation of the impact on reducing annotation costs was not accurate.
        \item The dependency of the model from the created dataset and vice versa.
        \item Other: \textit{[text input]}
    \end{enumerate}
    \item {[sc,m]} Would you consider applying Active Learning again in future annotation projects?
    \begin{enumerate}
        \item No
        \item Yes
        \item Other: \textit{[text input]}
    \end{enumerate}
    \item {[mc,c:V14]} Why would you not consider Active Learning in future annotation projects?
    \begin{enumerate}
        \item Active Learning has become obsolete nowadays.
        \item Active Learning is a useful concept in theory but does not work well enough in practice.
        \item Other: \textit{[text input]}
    \end{enumerate}
\end{enumerate}
}

\subsection*{\small VI. Trends in Active Learning for NLP}

{\small
In this section, we are interested in your opinion on whether the technological and methodological developments have affected the field of Active Learning, and what you think is still missing.\\

\noindent Please indicate how much you agree or disagree with the following statements:
\begin{enumerate}
    \item {[sc,m]} Text embeddings are crucial for many methods in Active Learning for NLP.
    \begin{itemize}
        \item Strongly agree, Agree, Neither agree nor disagree, Disagree, Strongly disagree \textit{(5-point Likert scale)}
    \end{itemize}
    \item {[sc,m]} GPU-accelerated computing has changed the choice of model in Active Learning for NLP.
    \begin{itemize}
        \item Strongly agree, Agree, Neither agree nor disagree, Disagree, Strongly disagree \textit{(5-point Likert scale)}
    \end{itemize}
    \item {[sc,m]} Neural Networks have changed the choice of instance selection strategy in Active Learning for NLP.
    \begin{itemize}
        \item Strongly agree, Agree, Neither agree nor disagree, Disagree, Strongly disagree \textit{(5-point Likert scale)}
    \end{itemize}
    \item {[sc,m]} LLMs (with 1B parameters or more) are frequently used in Active Learning for NLP.
    \begin{itemize}
        \item Strongly agree, Agree, Neither agree nor disagree, Disagree, Strongly disagree \textit{(5-point Likert scale)}\\
    \end{itemize}
    \item What are obvious next developments for Active Learning in NLP that you believe will have a big impact on the field? \textit{[text input]}
\end{enumerate}
}

\subsection*{\small VII. Background Information}

{\small
You are almost done. We just need a few more basic pieces of information about you in order to put your answers into context.

\begin{enumerate}
\item {[mc]} Where are you currently working? Please choose all options that apply to you.
\begin{enumerate}
    \item Academia 
    \item Industry
    \item Governmental organization
    \item Other: \textit{[text input]}
\end{enumerate}
\item {[mc]} What is your education background? Please choose all options that apply to you.
\begin{enumerate}
    \item Linguistics 
    \item Computational Linguistics 
    \item Computer Science/Informatics
    \item Engineering
    \item Mathematics
    \item Social Sciences
    \item Other: \textit{[text input]}
\end{enumerate}
\item {[mc]} Which field are you currently working in? Please choose all options that apply to you.
\begin{enumerate}
    \item Economics
    \item Education
    \item Financial sector
    \item Health services sector
    \item Humanities
    \item Information and communications technology
    \item Legal sector
    \item Public service sector
    \item Society and politics
    \item Other: \textit{[text input]}
\end{enumerate}
\item {[sc]} How many years of experience do you have in Machine Learning/NLP?
\begin{enumerate}
    \item Up to 1 year
    \item 1-2 years
    \item 3-5 years
    \item 5-10 years
    \item More than 10 years
\end{enumerate}
\item {[sc]} Which country is your primary place of residence?
\begin{enumerate}
    \item \textit{Dropdown menu with 196 country names}
    \item Other: \textit{[text input]}
\end{enumerate}
\item Do you have any comments on the survey? \textit{[text input]}
\item {[sc]} How did you become aware of the survey?
\begin{enumerate}
    \item Social media (e.g., Facebook, Twitter, Instagram)
    \item Personalized email invitation
    \item Mailing list (e.g., ACL Member Portal)
    \item Friend, colleague, or peer shared it
    \item Other: \textit{[text input]}
\end{enumerate}
\end{enumerate}
}

\normalsize

\section{Results}
\label{app:additional-results}

Table \ref{tab:responses-full} provides a full overview of survey participants' responses to the predefined options for all 52 questions. It is beyond the scope of this paper to cover all analyses enabled by the collected data, but the public release will allow researchers to conduct in-depth analyses of further aspects.

\subsection{Conservative Estimates (No Authors)}
\label{app:conservative-selection}

Our targeted recruitment of authors may have led to potential overestimation of AL in question groups I and II (the non-AL parts of the survey). To transparently account for potential author bias (response bias), our data includes information on recruitment methods (see~\question{VII.7}). By excluding the 33 participants recruited via personalized mails, we can provide a \textit{conservative baseline}. Excluding authors of AL publications provides the following picture:

\paragraph{RQ1} 106 out of 111 participants had encountered a lack of annotated data. With respect to potential changes due to recent advancements in NLP, participants largely agree that many problems will still require supervised learning (81\%; vs. 80\% in the full participant sample), and that annotated data remains a limiting factor; in general (93\%; vs. 94\% in the full sample), for certain languages (75\%; vs. 75\% in the full sample) and for problems of a certain complexity (92\%; vs. 91\% in the full sample). Fewer participants agree that generative AI for training data synthesis offers a viable solution (30\%; vs. 30\% in the full sample). \textbf{The response distribution of the conservative baseline closely mirrors that of the full participant sample.}

\paragraph{RQ2}
89 out of the 106 respondents had used computational methods.

\textbf{Adoption of active learning:}
AL was used by 36 respondents (40\%; vs.~53\% in the full sample). When asking why participants had not chosen AL, 25\% (vs.~25\% in the full sample) had never heard of it, and 51\% (vs.~49\% in the full sample) cited insufficient expertise. Methodological concerns cover the expected implementation overhead (40\%; vs.~37\% in the full sample), unsuitable annotation tools (34\%; vs.~32\% in the full sample), the difficulty of estimating effectiveness upfront (21\%; vs.~12\% in the full sample), and sampling bias (15\%; vs.~18\% in the full sample). 9\% (vs.~9\% in the full sample) found it unsuitable due to specific project requirements, and 26\% (vs.~14\% in the full sample) doubted AL’s effectiveness at all.

Despite these concerns, 54\% (vs.~54\% in the full sample) of those who had not yet used AL consider it for future projects and further 37\% (vs.~38\% in the full sample) indicate uncertainty due to a lack of knowledge. One participant would explicitly not use AL, four made its use dependent on the specific circumstances (similar to the full sample). \textbf{While response patterns are similar in general, the conservative sample emphasizes more the difficulty of upfront estimation of and general doubts about AL's effectiveness.}

\textbf{Alternative methods:}
The current users of AL also report obstacles in data annotation with AL. These are: finding annotators (36\%; vs.~35\% in the full sample), AL with multiple annotators (53\%; vs.~56\% in the full sample), unclear annotation instructions (31\%; vs.~29\% in the full sample), hard cases for annotation (53\%; vs.~59\% in the full sample), lack of GPU/TPU resources (22\%; vs.~25\% in the full sample), choice of model / query strategy (44\%; vs.~49\% in the full sample), choice of stopping criterion (36\%; vs.~35\% in the full sample), and other (6\%; vs.~6\% in the full sample). \textbf{The response distributions closely resemble each other.}

The use of computational alternatives to AL is widely dispersed across participants: data augmentation (70\%; vs.~73\% in the full sample), few-shot learning (57\%; vs.~63\% in the full sample), semi-supervised learning (51\%; vs.~53\% in the full sample), text generation (46\%; vs.~49\% in the full sample), transfer learning (62\% vs.~66\% in the full sample), weak supervision (40\%; vs.~44\% in the full sample), zero-shot learning (48\%; vs.~53\% in the full sample), and other (12\%; vs.~12\% in the full sample). \textbf{In the conservative baseline sample, AL is among the least used methods.} The alternative methods were considered successful in overcoming a lack of annotated data in slightly over half of the cases~(56\%; vs.~57\% in the full sample).

\subsection{Sector-specific Analysis of Section~\ref{sec:contemporary-active-learning}}
\label{app:group-analysis}

\begin{table*}[!bhpt]
\centering
\small
\begin{tabular}[!t]{@{}P{5cm}P{1cm}P{1cm}P{1cm}P{1cm}P{1cm}P{1cm}@{}}
\toprule
\textbf{\scshape Variable} & \multicolumn{3}{c}{\textbf{\scshape {Academia}}} & \multicolumn{3}{c}{\textbf{\scshape {Industry/Government}}}\\\
& \# no & \# yes & $\mu_{\textnormal{yes}}$ &  \# no & \# yes & $\mu_{\textnormal{yes}}$ \\
\midrule
Insufficient expertise (a) & 9 & 2 & 0.18& 14   & 1& 0.07 \\
Overhead in setup (b) & 7 &  4  & 0.36 & 11 & 4 & 0.27 \\
Lack of annotation tools (c)& 6 & 5 & 0.45 & 9 & 6& 0.4 \\
Project-specific requirements (d) & 10 & 1 & 0.09 & 14 & 1& 0.07 \\
Sampling bias (e)   & 8 & 3  & 0.27& 12 & 3& 0.20 \\
No cost reduction (f)& 11& 0  & 0.00 & 12  & 3& 0.20  \\
Did not work well (g)& 7 & 4   & 0.36& 11 & 4  & 0.27\\
Inaccurate upfront estimation (h)& 10 & 1 & 0.09& 12 & 3 & 0.2 \\
Dataset-model dependency (i)& 7 & 4  & 0.36 & 13 & 2& 0.13 \\
\bottomrule
\end{tabular}
\caption{
Predefined reason selection for reduced effectiveness in AL (\protect\question{V.13}) in practical projects (2020 onward) across sectors. We show the frequency of no (\# no; the respective reason did not affect the effectiveness of AL) and yes answers (\# yes; the respective reason affected the effectiveness of AL), and the mean of yes answers ($\mu_\textnormal{yes}$).
}
\label{table-reduced-effectiveness}
\end{table*}

In Section \ref{sec:contemporary-active-learning}, we assess the practical utility of contemporary AL by including respondents from academia, industry and governmental organizations. This way, we do not make a distinction between experiences reported from NLP researchers in academic applications and practical utility ``in the real world'', as we believe that using AL-supported annotation for the creation of research data also offers valid practical insights (in contrast to the AL-simulation lab scenario without human annotators). Nonetheless, there may be considerable differences in usage that should be analyzed further to inform future work. What follows, therefore, presents a distinction between academic (16) and industrial/governmental respondents (20, with 3 overlaps) based on the 33 recent projects.

\paragraph{Models, query strategies, and stopping criteria}
Both groups show similar preferences for LLMs, with about two-thirds in each group using them, primarily the smaller models. Industry/governmental participants rely more on uncertainty sampling (50\% vs. 25\%), and their waiting times vary more---possibly due to a larger data~pool.

\paragraph{Annotation tools for active learning}
Annotation tool use differs notably: 70\% of industry/governmental participants use tools vs. only 23\% of those that exclusively work in academia. 

\paragraph{Satisfaction with project}
The answer distributions on the Likert scales for success and effectiveness differed slightly between industry/government and academia. However, the main trend is clear: 85\% respectively 94\% found their projects (very) successful, and 65\% respectively 63\% found them (very) effective. In case of the 23 cases of reduced effectiveness, Table~\ref{table-reduced-effectiveness} provides an overview of the response patterns for reasons. Overall, both groups exhibit similar tendencies in the yes/no responses. Especially, for the lack of suitable annotation tools (c), unmet project-specific requirements (d), and sampling bias (e), the mean values in the sample at hand are closely aligned. Slightly more difference can be observed in an inaccurate upfront estimation (h) or no reduction (f) of annotation costs, with no academic respondents indicating the latter issue. In contrast, model-dependency of datasets (i) seems to be more of a concern in academia, as is insufficient expertise (a).\footnote{We refrain from applying significance tests due to the small sample size, which limits the generalizability of the observations and, thus, the informative value of such analysis.} 90\% of industry/governmental respondents would use AL again without restrictions, compared to 81\% of academic respondents, who are slightly more critical.

To sum up, while there are \textit{some usage differences}, \textit{both groups align closely on effectiveness and reasons for reduced effectiveness}, with only minor variations. Our findings support the identified main challenges for contemporary AL across NLP researchers and ``real-world'' applicants.

\begin{figure}[t]
\centering
\includegraphics[width=0.41\textwidth]{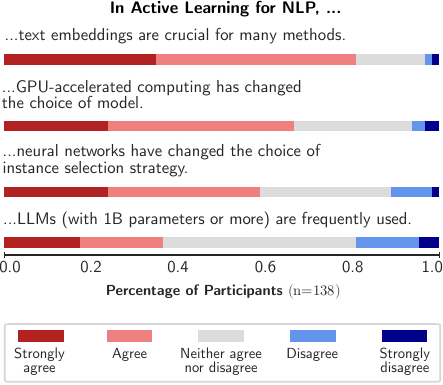}
\caption{Participants' assessment of the impact of various methodological and technological developments on AL using a 5-point Likert scale (cf.~\protect\questionlink{VI.1}{VI.1--4}).}
\label{likert-trends-in-active-learning}
\end{figure}

\subsection{Selected Additional Analysis}

\paragraph{Order of annotated data}
The free-text answers to \question{V.4} inform us about the volume of annotated data collected in the reported annotation projects. For the 33 recent applied AL projects, we observe a range from 50 to 1,000,000 annotation units, with an average of 43,265 (three entries miss size information). Examining differences between industry and academic respondents, we find that the former report creating substantially less annotated data (on average 2,313 units) than the latter (on average 86,506 units). This may reflect more restrictive requirements in industrial applications.

\paragraph{Impact of recent developments in NLP on AL}

Table \ref{likert-trends-in-active-learning} illustrates the perception of how recent developments in NLP have impacted AL.

\onecolumn

\input{table-responses}

\twocolumn

%% file: table-responses.tex
{
\footnotesize
\renewcommand{\arraystretch}{1.3}
\begin{longtable}{@{}P{1.6cm}P{1.5cm}P{0.75cm}P{10.5cm}@{}}
\toprule 
\textbf{\scshape Survey Question} & \textbf{\scshape Mapping to RQ} & $n$ & \textbf{\scshape Responses (total / percentage of participants)}\\ 
\midrule
\endfirsthead
\multicolumn{4}{c}%
{{\bfseries \tablename\ \thetable{} -- continued from previous page}} \\
\midrule 
\textbf{\scshape Survey Question} & \textbf{\scshape Mapping to RQ} & $n$ & \textbf{\scshape Responses (total / percentage of participants)}\\ \hline 
\endhead
\midrule 
\multicolumn{4}{r}{{Continued on next page}} \\ 
\midrule
\endfoot
\bottomrule \\[-2.5ex]
\caption{Overview of response selection. Each survey question is linked to a research question. We indicate the number of participants $n$ per question and their distribution across predefined response options. In case of questions described in Section \ref{sec:contemporary-active-learning}, we provide both the full set of respondents ($n=48$) and the set of respondents that started the reported AL project in 2020 or later ($n=33$, marked in bold). Free-text responses are omitted for scope but can be seen in the dataset, provided in the supplementary material. \textasteriskcentered{}: marks questions not discussed in the main body of the paper, \textasteriskcentered\textasteriskcentered{}: marks  questions only discussed partly.\\
}
\label{tab:responses-full} 
\endlastfoot
\qlabel{I.1}\bf I.1 & RQ1 & 144 & Yes (138 / 96\%), No (6 / 4\%)\\
\qlabel{I.2}\bf I.2 & RQ1 & 138 & When working on following under-resourced languages (77 / 56\%), When working on a certain task (96 / 70\%), When working with a specific requirement~(71~/~51\%), Other~(119~/~86\%)\\
\qlabel{I.3}\bf I.3 & RQ1 & 138 & Strongly agree (3 / 2\%), Agree (38 / 28\%), Neither agree nor disagree~(38~/~28\%), Disagree (39 / 28\%), Strongly disagree (20 / 14\%)\\
\qlabel{I.4}\bf I.4 & RQ1 & 138 & Strongly agree (35 / 25\%), Agree (75 / 54\%), Neither agree nor disagree~(18~/~13\%), Disagree (8 / 6\%), Strongly disagree (2 / 1\%)\\
\qlabel{I.5}\bf I.5 & RQ1 & 138 & Strongly agree (59 / 43\%), Agree (71 / 51\%), Neither agree nor disagree~(6~/~4\%), Disagree (1 / 1\%), Strongly disagree (1 / 1\%)\\
\qlabel{I.6}\bf I.6 & RQ1 & 138 & Strongly agree (51 / 37\%), Agree (53 / 38\%), Neither agree nor disagree~(25~/~18\%), Disagree (8 / 6\%), Strongly disagree (1 / 1\%)\\
\qlabel{I.7}\bf I.7 & RQ1 & 138 & Strongly agree (62 / 45\%), Agree (63 / 46\%), Neither agree nor disagree~(11~/~8\%), Disagree (1 / 1\%), Strongly disagree (1 / 1\%)\\
\midrule
\qlabel{II.1}\bf II.1 & RQ2 & 138 & Yes (120 / 87\%), No (18 / 13\%) \\
\qlabel{II.2}\bf II.2 & RQ2 & 120 & Active Learning (63 / 53\%), Data augmentation (88 / 73\%), Few-shot Learning~(75~/~63\%), Semi-supervised Learning (64 / 53\%), Text generation~(59~/~49\%), Transfer learning (79 / 66\%), Weak supervision (53 / 44\%), Zero-shot (63 / 53\%), Other~(14~/~12\%)\\
\qlabel{II.3}\bf II.3\textasteriskcentered & RQ2 & 18 & I have never heard of these methods. (10 / 56\%), I never coordinated an annotation project. (6 / 33\%), We had enough human resources to annotate our data~(5~/~28\%).\\
\qlabel{II.4}\bf II.4 & RQ2 & 56\footnote{One participant was not directed to this question after choosing solely ``Other'' in II.2. This applies also for II.7.} & Strongly agree (4 / 7\%), Agree (28 / 50\%), Neither agree nor disagree (21 / 38\%), Disagree (2 / 4\%), Strongly disagree (1 / 2\%)\\
\qlabel{II.5}\bf II.5 & RQ2, RQ5 & 57 & Never heard of Active Learning. (14 / 25\%), Insufficient expertise / knowledge.~(28~/~49\%), Did not want to spend overhead in implementing an Active Learning-based annotation environment. (21 / 37\%), Did not know of any suitable annotation tools that easily integrate Active Learning. (18 / 32\%), Did not meet my project's specific requirements. (5 / 9\%), Wanted to avoid sampling bias in the corpus.~(10~/~18\%), Was not convinced that Active Learning would reduce annotation cost.~(14~/~25\%), Did not use Active Learning because I couldn't estimate upfront its impact on reducing annotation costs. (12 / 21\%), Other (9 / 16\%)\\
\qlabel{II.6}\bf II.6\textasteriskcentered & RQ2, RQ5 & 5 & \textit{free text answer} (5 / 100\%)\\
\qlabel{II.7}\bf II.7 & RQ2, RQ5 & 56 &  Yes (30 / 54\%), No (1 / 2\%), I don’t know enough about Active Learning to answer this question. (21 / 38\%), Only under specific circumstances (4 / 7\%)\\
\qlabel{II.8}\bf II.8 & RQ2 & 1 & Active Learning has become obsolete nowadays. (0 / 0\%), Active Learning is a useful concept in theory but does not work well enough in practice. (1 / 100\%), Other~(0~/~0\%)\\
\midrule
\qlabel{III.1}\bf III.1\textasteriskcentered & RQ2, RQ3 & 63 & I have been part of an Active Learning workflow as an annotator. (24 / 38\%), I have organized an Active Learning annotation workflow. (38 / 60\%), I have knowledge about instance selection strategies in Active Learning but use pre-built solutions.~(28~/~44\%), I have implemented Active Learning strategies or workflows.~(45~/~71\%), I have researched Active Learning. (43 / 68\%)\\
\qlabel{III.2}\bf III.2 & RQ2, RQ5 & 63 & Obtain annotated data at minimal annotation costs (55 / 87\%), Gather experience with practical applications of Active Learning (29 / 46\%), Test your annotation guidelines to identify difficult examples (15 / 24\%), Other (5 / 8\%)\\
\qlabel{III.3}\bf III.3 & RQ2 & 63 & Finding annotators that achieve satisfactory performance (22 / 35\%), Performing Active Learning with multiple annotators (35 / 56\%), Unclear instructions how to annotate~(18~/~29\%), Hard cases for annotation (37 / 59\%), Lack of resources~(16~/~25\%), Choice of model and instance selection algorithm (31 / 49\%), Choice of stopping criterion (22 / 35\%), Other (4 / 6\%)\\
\qlabel{III.4}\bf III.4\textasteriskcentered\textasteriskcentered & RQ3 & 63 & Less than a minute (6 / 10\%), 1 minute to less than 10 minutes (13 / 21\%), 10 minutes to less than 1 hour (3 / 5\%), 1 hour to less than 10 hours (9 / 14\%), 10 hours to less than 1 day (8 / 13\%), 1 day to less than 1 week (11 / 17\%), I will wait long as it takes~(13~/~21\%)\\
\qlabel{III.5}\bf III.5 & RQ3 & 63 & In a lab setting with simulated annotators (36 / 57\%), In an applied setting with human annotators (48 / 76\%)\\
\midrule
\qlabel{IV.1}\bf
IV.1\textasteriskcentered\textasteriskcentered\footnote{The years of project start in the free text answer are: 2020--2024 (33), 2015--2019 (11), 2010--2014 (2), 2005--2009 (1).} & RQ3 & 48 & \textit{free text answer} (47 / 98\%), No answer (1 / 2\%)\\
\qlabel{IV.2}\bf IV.2\footnote{For the paper, we refined the wording by replacing ``classic models'' with ``traditional models''.} & RQ3, RQ5 & 48 & Classic (11 / 23\%), Neural Network (5 / 8\%), BERT era Language Models such as BERT, GPT-2, RoBERTa, ... (24 / 50\%), Large Language Models such as \mbox{Phi-3}, Llama, Mistral, Gemma, Gemini, GPT-3/4, ... (2 / 4\%), Do not know (2 / 4\%), Other~(4~/~8\%)\\
&& \textbf{33} & Classic (3 / 9\%), Neural Network (4 / 12\%), BERT era Language Models such as BERT, GPT-2, RoBERTa, ... (20 / 61\%), Large Language Models such as \mbox{Phi-3}, Llama, Mistral, Gemma, Gemini, GPT-3/4, ... (2 / 6\%), Do not know (2 / 6\%), Other~(2~/~6\%)\\
\qlabel{IV.3}\bf IV.3 & RQ3, RQ5 & 48 & Uncertainty-based (20 / 42\%), Disagreement-based (6 / 13\%), Gradient information-based (2 / 4\%), Performance prediction-based (5 / 10\%), Density-based (1 / 2\%), Discriminative-based (0 / 0\%), Batch diversity (1 / 2\%), Hybrid~(7~/~15\%), Do not know~(4~/~8\%), Other (2 / 4\%)\\
& & \textbf{33} & Uncertainty-based (13 / 39\%), Disagreement-based (5 / 15\%), Gradient information-based (2 / 6\%), Performance prediction-based (4 / 12\%), Density-based (1 / 3\%), Discriminative-based (0 / 0\%), Batch diversity (1 / 3\%), Hybrid~(3~/~9\%), Do not know~(3~/~9\%), Other (1 / 3\%)\\
\qlabel{IV.4}\bf IV.4\textasteriskcentered\textasteriskcentered & RQ3 & 48 & We did not retrain (0 / 0\%), Less than a minute (8 / 17\%), 1 minute to less than 10 minutes (8 / 17\%), 10 minutes to less than 1 hour (6 / 13\%), 1 hour to less than 10 hours~(8~/~17\%), 10 hours to less than 1 day (6 / 13\%), 1 day to less than 1 week~(6~/~13\%), Do not know (6 / 13\%)\\
&& \bf 33 & We did not retrain (0 / 0\%), Less than a minute (3 / 10\%), 1 minute to less than 10 minutes (6 / 21\%), 10 minutes to less than 1 hour (6 / 21\%), 1 hour to less than 10 hours~(4~/~14\%), 10 hours to less than 1 day (5 / 17\%), 1 day to less than 1 week~(5~/~17\%), Do not know (4 / 0\%)\\
\qlabel{IV.5}\bf IV.5 & RQ3, RQ5 & 48 & Evaluation of the learned model on a held-out gold standard (15 / 31\%), Money and/or time available for annotation were depleted (18 / 38\%), All relevant documents annotated~(4~/~8\%), The stopping time was indicated by an algorithmic stopping criterion~(6~/~13\%), Do not know (2 / 4\%), Other (3 / 6\%)\\
&  & \bf 33 & Evaluation of the learned model on a held-out gold standard (10 / 30\%), Money and/or time available for annotation were depleted (13 / 39\%), All relevant documents annotated~(2~/~6\%), The stopping time was indicated by an algorithmic stopping criterion~(4~/~12\%), Do not know (1 / 3\%), Other (3 / 9\%)\\
\midrule
\qlabel{V.1}\bf V.1\textasteriskcentered\footnote{We updated the 2009 task selection \cite{tomanek-olsson-2009-web} based on \href{https://web.archive.org/web/20240518153154/http://nlpexplorer.org/}{http://nlpexplorer.org/} and \citet{hou-etal-2021-tdmsci}.} & RQ3, RQ5 & 48 & Automatic speech recognition (0 / 0\%), Coreference resolution (1 / 2\%), Chunking~(0~/~0\%), Information extraction (5 / 10\%), Language generation (0 / 0\%), Language understanding ((0 / 0\%)), Machine translation (3 / 6\%), Morphological analysis~(0~/~0\%), Named entity recognition (5 / 10\%), Part-of-speech tagging~(0~/~0\%), Question answering~(0~/~0\%), Relation extraction~(1~/~2\%), Semantic similarity~(0~/~0\%), Sentiment analysis (3 / 6\%), Syntactic parsing (1 / 2\%), Summarization~(0~/~0\%), Text categorization~(22~/~46\%), Word segmentation (0 / 0\%), Word sense disambiguation (0 / 0\%), Other~(7~/~15\%)\\
\qlabel{V.2}\bf V.2\textasteriskcentered\footnote{We adopted the language selection for this question from \citet{tomanek-olsson-2009-web}.} & RQ3, RQ5 & 48 & Arabic (4 / 8\%), English (32 / 67\%), French (5 / 10\%), German (10 / 21\%), Hindi~(1~/~1\%), Mandarin (1 / 2\%), Spanish (6 / 13\%), Other (16 / 33\%)\\
\qlabel{V.3}\bf V.3\textasteriskcentered & RQ3, RQ5 & 48 & \textit{Indication of hours} (22 / 46\%), Do not know (26 / 54\%)\\
\qlabel{V.4}\bf V.4\textasteriskcentered & RQ3, RQ5 & 48 & \textit{free text answer} (48 / 100\%)\\
\qlabel{V.4}\bf V.5\textasteriskcentered & RQ3 & 48 & Annotation was done by the project coordinators or other project members~(20~/~42\%), Domain experts (29 / 60\%), Non-domain experts (12 / 25\%), Other (3 \ 6\%)\\
\qlabel{V.6}\bf V.6 & RQ3 & 48 & Yes (25 / 52\%), No (23 / 48\%)\\
&& \bf 33 & Yes (17 / 52\%), No (16 / 48\%)\\
\qlabel{V.7}\bf V.7\footnote{We based the list of annotation tools on \citet{borisova:2024}. For $n=48$, in addition to 7 self-built solutions, participant-added options emphasize further relevant tools: LabelSleuth (4; \citealp{shnarch:2022}), AL Toolbox (2; \citealp{tsvigun:2022}), ALAMBIC (1; \citealp{nachtegael:2023}), and ALANNO (1; \citealp{jukic:2023}). For $n=33$, in addition to 4 self-built solutions, participant-added options emphasize further relevant tools: LabelSleuth (3), ALAMBIC (1), and ALANNO (1).} & RQ3 & 25 & ActiveAnno (1 / 4\%), Argilla (4 / 16\%), AWS Sagemaker Ground Truth~(0~/~0\%), AWS Comprehend (0 / 0\%), BioQRator (0 / 0\%), INCEpTION (2 / 8\%), Labelbox~(0~/~0\%), Label Studio (1 / 4\%), MAT (0 / 0\%), Potato (0 / 0\%), Prodigy~(2~/~8\%), Other~(15~/~31\%)\\
&& \bf 17 & ActiveAnno (1 / 6\%), Argilla (4 / 24\%), AWS Sagemaker Ground Truth~(0~/~0\%), AWS Comprehend (0 / 0\%), BioQRator (0 / 0\%), INCEpTION (0 / 0\%), Labelbox~(0~/~0\%), Label Studio (1 / 6\%), MAT (0 / 0\%), Potato (0 / 0\%), Prodigy~(2~/~12\%), Other~(9~/~53\%)\\
\qlabel{V.8}\bf V.8\textasteriskcentered & RQ3 & 48 & Deleted them (6 / 13\%), Assigned one of two labels (14 / 29\%), Not applicable, each instance had at most one annotation. (15 / 31\%), Other (13 / 27\%)\\
\qlabel{V.9}\bf V.9\textasteriskcentered & RQ3 & 48 & No (27 / 56\%), Yes (21 / 44\%)\\
\qlabel{V.10}\bf V.10\textasteriskcentered & RQ3 & 21 & Previously annotated examples were re-annotated. (12 / 57\%), Previously annotated examples were left unchanged. (6 / 29\%), Other (3 / 14\%)\\
\qlabel{V.11}\bf V.11 & RQ3 & 48 & Very successful (7 / 15\%), Successful (35 / 73\%), Neither successful nor unsuccessful~(6~/~13\%), Unsuccessful (0 / 0\%), Very unsuccessful (0 / 0\%)\\
&& \bf 33 & Very successful (6 / 18\%), Successful (24 / 73\%), Neither successful nor unsuccessful~(3~/~9\%), Unsuccessful (0 / 0\%), Very unsuccessful (0 / 0\%)\\
\qlabel{V.12}\bf V.12 & RQ3, RQ5 & 48 & Very effective (12 / 25\%), Effective (19 / 40\%), Neither effective nor ineffective~(14~/~29\%), Ineffective (2 / 4\%), Very ineffective (1 / 2\%)\\
&& \bf 33 & Very effective (10 / 30\%), Effective (12 / 36\%), Neither effective nor ineffective~(8~/~24\%), Ineffective (2 / 6\%), Very ineffective (1 / 3\%)\\
\qlabel{V.13}\bf V.13 & RQ3, RQ5 & 36 & In retrospect, I had insufficient expertise/knowledge. (6 / 17\%), Overhead in setting up an Active Learning-based annotation environment. (10 / 28\%), Lack of suitable annotation tools that easily integrate Active Learning. (11 / 31\%), In retrospect, Active Learning did not meet my project's specific requirements.~(2~/~6\%), Active Learning created a sampling bias in the corpus. (6 / 17\%), Active Learning did not reduce annotation cost.~(4~/~11\%), Active Learning did not work well in my scenario.~(7~/~19\%), My upfront estimation of the impact on reducing annotation costs was not accurate.~(5~/~14\%), The dependency of the model from the created dataset and vice versa.~(7~/~19\%), Other~(8~/~22\%)\\
&& \bf 23 & In retrospect, I had insufficient expertise/knowledge. (3 / 9\%), Overhead in setting up an Active Learning-based annotation environment. (7 / 21\%), Lack of suitable annotation tools that easily integrate Active Learning. (8 / 24\%), In retrospect, Active Learning did not meet my project's specific requirements.~(2~/~6\%), Active Learning created a sampling bias in the corpus. (5 / 15\%), Active Learning did not reduce annotation cost.~(3~/~9\%), Active Learning did not work well in my scenario.~(7~/~21\%), My upfront estimation of the impact on reducing annotation costs was not accurate.~(3~/~9\%), The dependency of the model from the created dataset and vice versa.~(5~/~15\%), Other~(4~/~12\%)\\
\qlabel{V.14}\bf V.14 & RQ3, RQ5 & 48 & No (2 / 4\%), Yes (42 / 88\%), Other (4 / 8\%)\\
&& \bf 33 & No (2 / 6\%), Yes (28 / 85\%), Other (3 / 9\%)\\
\qlabel{V.15}\bf V.15 & RQ3, RQ5 & 2 & Active Learning has become obsolete nowadays. (0 / 0\%), Active Learning is a useful concept in theory but does not work well enough in practice. (2 / 100\%), Other~(0~/~0\%)\\
&& \bf 2 & Active Learning has become obsolete nowadays. (0 / 0\%), Active Learning is a useful concept in theory but does not work well enough in practice. (2 / 100\%), Other~(0~/~0\%)\\
\midrule
\qlabel{VI.1}\bf VI.1 & RQ4 & 63 & Strongly agree (22 / 35\%), Agree (29 / 46\%), Neither agree nor disagree~(10~/~16\%), Disagree (1 / 2\%), Strongly disagree (1 / 2\%)\\
\qlabel{VI.2}\bf VI.2 & RQ4 & 63 & Strongly agree (15 / 24\%), Agree (27 / 43\%), Neither agree nor disagree~(17~/~27\%), Disagree (2 / 3\%), Strongly disagree (2 / 3\%)\\
\qlabel{VI.3}\bf VI.3 & RQ4 & 63 & Strongly agree (15 / 24\%), Agree (22 / 35\%), Neither agree nor disagree~(19~/~30\%), Disagree (6 / 10\%), Strongly disagree (1 / 2\%)\\
\qlabel{VI.4}\bf VI.4 & RQ4 & 63 & Strongly agree (11 / 17\%), Agree (12 / 19\%), Neither agree nor disagree~(28~/~44\%), Disagree (9 / 14\%), Strongly disagree (3 / 5\%)\\
\qlabel{VI.5}\bf VI.5 & RQ4 & 63 & \textit{free text answer} (26 / 41\%), No answer (37 / 59\%)\\
\midrule
\qlabel{VII.1}\bf VII.1 & background & 144 & Academia (107 / 74\%), Industry (45 / 31\%), Governmental organization (7 / 5\%), Other~(2~/~1\%)\\
\qlabel{VII.2}\bf VII.2 & background & 144 & Linguistics (27 / 19\%), Computational Linguistics (56 / 39\%), Computer Science/Informatics (88 / 61\%), Engineering (23 / 16\%), Mathematics (10 / 7\%), Social Sciences (14 / 10\%), Other (6 / 4\%)\\
\qlabel{VII.3}\bf VII.3 & background & 144 & Economics (4 / 3\%), Education (44 / 31\%), Financial sector (5 / 3\%), Health services sector (12 / 8\%), Humanities (24 / 17\%), Information and communications technology~(79~/~55\%), Legal sector (8 / 6\%), Public service sector (11 / 8\%), Society and politics (7 / 5\%), Other (18 / 13\%)\\
\qlabel{VII.4}\bf VII.4 & background & 144 & Up to 1 year (1 / 1\%), 1-2 years (11 / 8\%), 3-5 years (47 / 33\%), 5-10 years~(55~/~38\%), More than 10 years (30 / 21\%)\\
\qlabel{VII.5}\bf VII.5 & background & 144 & \textit{Indication of country} (143 / 99\%), Other (0 / 0\%), No answer (1 / 1\%)\\
\qlabel{VII.6}\bf VII.6\textasteriskcentered & background & 144 & \textit{free-text answer} (22 / 15\%), No answer (122 / 85\%)\\
\qlabel{VII.7}\bf VII.7\textasteriskcentered\textasteriskcentered & background & 144 & Social media (16 / 11\%), Personalized email invitation (33 / 23\%), Mailing list~(73~/~51\%), Friend, colleague, or peer shared it (13 / 9\%), Other (6 / 4\%), No answer (3 / 2\%)\\
\end{longtable}}

%% file: main.bbl
\begin{thebibliography}{61}
\providecommand{\natexlab}[1]{#1}

\bibitem[{Baumann et~al.(2025)Baumann, Röttger, Urman, Wendsjö, del Arco,
  Gruber, and Hovy}]{baumann:2025}
Joachim Baumann, Paul Röttger, Aleksandra Urman, Albert Wendsjö, Flor
  Miriam~Plaza del Arco, Johannes~B. Gruber, and Dirk Hovy. 2025.
\newblock \href {https://arxiv.org/abs/2509.08825} {Large language model
  hacking: Quantifying the hidden risks of using llms for text annotation}.
\newblock \emph{Preprint}, arXiv:2509.08825.

\bibitem[{Bayer and Reuter(2024)}]{bayer:2024}
Markus Bayer and Christian Reuter. 2024.
\newblock \href {https://arxiv.org/abs/2405.10808} {Activellm: Large language
  model-based active learning for textual few-shot scenarios}.
\newblock \emph{Preprint}, arXiv:2405.10808.

\bibitem[{Blaschke et~al.(2024)Blaschke, Purschke, Schuetze, and
  Plank}]{blaschke:2024}
Verena Blaschke, Christoph Purschke, Hinrich Schuetze, and Barbara Plank. 2024.
\newblock \href {https://doi.org/10.18653/v1/2024.acl-short.74} {What do
  dialect speakers want? a survey of attitudes towards language technology for
  {G}erman dialects}.
\newblock In \emph{Proceedings of the 62nd Annual Meeting of the Association
  for Computational Linguistics (Volume 2: Short Papers)}, pages 823--841,
  Bangkok, Thailand. Association for Computational Linguistics.

\bibitem[{Borisova et~al.(2024)Borisova, Abu~Ahmad, Garcia-Castro, Usbeck, and
  Rehm}]{borisova:2024}
Ekaterina Borisova, Raia Abu~Ahmad, Leyla Garcia-Castro, Ricardo Usbeck, and
  Georg Rehm. 2024.
\newblock \href {https://aclanthology.org/2024.law-1.4/} {Surveying the
  {FAIR}ness of annotation tools: Difficult to find, difficult to reuse}.
\newblock In \emph{Proceedings of The 18th Linguistic Annotation Workshop
  (LAW-XVIII)}, pages 29--45, St. Julians, Malta. Association for Computational
  Linguistics.

\bibitem[{Ein-Dor et~al.(2020)Ein-Dor, Halfon, Gera, Shnarch, Dankin, Choshen,
  Danilevsky, Aharonov, Katz, and Slonim}]{ein-dor:2020}
Liat Ein-Dor, Alon Halfon, Ariel Gera, Eyal Shnarch, Lena Dankin, Leshem
  Choshen, Marina Danilevsky, Ranit Aharonov, Yoav Katz, and Noam Slonim. 2020.
\newblock \href {https://doi.org/10.18653/v1/2020.emnlp-main.638} {{A}ctive
  {L}earning for {BERT}: {A}n {E}mpirical {S}tudy}.
\newblock In \emph{Proceedings of the 2020 Conference on Empirical Methods in
  Natural Language Processing (EMNLP)}, pages 7949--7962, Online. Association
  for Computational Linguistics.

\bibitem[{Galimzianova and Sanochkin(2024)}]{galimzianova:2024}
Daria Galimzianova and Leonid Sanochkin. 2024.
\newblock \href {https://doi.org/10.18653/v1/2024.findings-emnlp.840}
  {Efficient active learning with adapters}.
\newblock In \emph{Findings of the Association for Computational Linguistics:
  EMNLP 2024}, pages 14374--14383, Miami, Florida, USA. Association for
  Computational Linguistics.

\bibitem[{Ghose and Nguyen(2024)}]{ghose-nguyen-2024-fragility}
Abhishek Ghose and Emma~Thuong Nguyen. 2024.
\newblock \href {https://doi.org/10.18653/v1/2024.emnlp-main.1240} {On the
  fragility of active learners for text classification}.
\newblock In \emph{Proceedings of the 2024 Conference on Empirical Methods in
  Natural Language Processing}, pages 22217--22233, Miami, Florida, USA.
  Association for Computational Linguistics.

\bibitem[{H{\o}jer et~al.(2025)H{\o}jer, Thorn~Jakobsen, Rogers, and
  Heinrich}]{hojer-etal-2025-research}
Bertram H{\o}jer, Terne~Sasha Thorn~Jakobsen, Anna Rogers, and Stefan Heinrich.
  2025.
\newblock \href {https://doi.org/10.18653/v1/2025.findings-acl.1324} {Research
  community perspectives on ``intelligence'' and large language models}.
\newblock In \emph{Findings of the Association for Computational Linguistics:
  ACL 2025}, pages 25796--25812, Vienna, Austria. Association for Computational
  Linguistics.

\bibitem[{Hou et~al.(2021)Hou, Jochim, Gleize, Bonin, and
  Ganguly}]{hou-etal-2021-tdmsci}
Yufang Hou, Charles Jochim, Martin Gleize, Francesca Bonin, and Debasis
  Ganguly. 2021.
\newblock \href {https://doi.org/10.18653/v1/2021.eacl-main.59} {{TDMS}ci: A
  specialized corpus for scientific literature entity tagging of tasks datasets
  and metrics}.
\newblock In \emph{Proceedings of the 16th Conference of the European Chapter
  of the Association for Computational Linguistics: Main Volume}, pages
  707--714, Online. Association for Computational Linguistics.

\bibitem[{Jeleni{\'c} et~al.(2023)Jeleni{\'c}, Juki{\'c}, Drobac, and
  Snajder}]{jelenic-etal-2023-dataset}
Fran Jeleni{\'c}, Josip Juki{\'c}, Nina Drobac, and Jan Snajder. 2023.
\newblock \href {https://doi.org/10.18653/v1/2023.findings-acl.144} {On dataset
  transferability in active learning for transformers}.
\newblock In \emph{Findings of the Association for Computational Linguistics:
  ACL 2023}, pages 2282--2295, Toronto, Canada. Association for Computational
  Linguistics.

\bibitem[{Juki{\'c} et~al.(2023)Juki{\'c}, Jeleni{\'c}, Bi{\'c}ani{\'c}, and
  Snajder}]{jukic:2023}
Josip Juki{\'c}, Fran Jeleni{\'c}, Miroslav Bi{\'c}ani{\'c}, and Jan Snajder.
  2023.
\newblock \href {https://doi.org/10.18653/v1/2023.eacl-demo.26} {{ALANNO}: An
  active learning annotation system for mortals}.
\newblock In \emph{Proceedings of the 17th Conference of the European Chapter
  of the Association for Computational Linguistics: System Demonstrations},
  pages 228--235, Dubrovnik, Croatia. Association for Computational
  Linguistics.

\bibitem[{Kholodna et~al.(2024)Kholodna, Julka, Khodadadi, Gumus, and
  Granitzer}]{kholodna:2024}
Nataliia Kholodna, Sahib Julka, Mohammad Khodadadi, Muhammed~Nurullah Gumus,
  and Michael Granitzer. 2024.
\newblock \href {https://doi.org/10.1007/978-3-031-70381-2_25} {Llms in the
  loop: Leveraging large language model annotations for active learning in
  low-resource languages}.
\newblock In \emph{Machine Learning and Knowledge Discovery in Databases.
  Applied Data Science Track: European Conference, ECML PKDD 2024, Vilnius,
  Lithuania, September 9–13, 2024, Proceedings, Part X}, page 397–412,
  Berlin, Heidelberg. Springer-Verlag.

\bibitem[{Klie et~al.(2018)Klie, Bugert, Boullosa, Eckart~de Castilho, and
  Gurevych}]{klie-etal-2018-inception}
Jan-Christoph Klie, Michael Bugert, Beto Boullosa, Richard Eckart~de Castilho,
  and Iryna Gurevych. 2018.
\newblock \href {https://aclanthology.org/C18-2002/} {The {INCE}p{TION}
  platform: Machine-assisted and knowledge-oriented interactive annotation}.
\newblock In \emph{Proceedings of the 27th International Conference on
  Computational Linguistics: System Demonstrations}, pages 5--9, Santa Fe, New
  Mexico. Association for Computational Linguistics.

\bibitem[{Kohl et~al.(2024)Kohl, Kr{\"a}mer, Fohry, and Kraft}]{kohl2024er}
Philipp Kohl, Yoka Kr{\"a}mer, Claudia Fohry, and Bodo Kraft. 2024.
\newblock \href {https://doi.org/10.1007/978-3-031-66694-0_6} {Scoping review
  of active learning strategies and their evaluation environments for entity
  recognition tasks}.
\newblock In \emph{Deep Learning Theory and Applications}, pages 84--106, Cham.
  Springer Nature Switzerland.

\bibitem[{Kwon et~al.(2013)Kwon, Kim, Shin, and Wilbur}]{Kwon2013BioQRatorA}
Dongseop Kwon, Sun Kim, Soo~Yong Shin, and John Wilbur. 2013.
\newblock Bioqrator: a web-based interactive biomedical literature curating
  system.
\newblock In \emph{Proceedings of the fourth biocreative challenge evaluation
  workshop}, volume~1, pages 241--246.

\bibitem[{Lewis and Gale(1994)}]{lewis:1994}
David~D. Lewis and William~A. Gale. 1994.
\newblock \href {https://doi.org/10.1007/978-1-4471-2099-5\_1} {A sequential
  algorithm for training text classifiers}.
\newblock In \emph{Proceedings of the 17th Annual International {ACM-SIGIR}
  Conference on Research and Development in Information Retrieval}, pages
  3--12. Springer, ACM/Springer.

\bibitem[{Li et~al.(2024)Li, Zhang, Wang, Tan, Kosugi, and Okumura}]{li:2024}
Dongyuan Li, Ying Zhang, Zhen Wang, Shiyin Tan, Satoshi Kosugi, and Manabu
  Okumura. 2024.
\newblock \href {https://doi.org/10.18653/v1/2024.findings-emnlp.523} {Active
  learning for abstractive text summarization via {LLM}-determined curriculum
  and certainty gain maximization}.
\newblock In \emph{Findings of the Association for Computational Linguistics:
  EMNLP 2024}, pages 8959--8971, Miami, Florida, USA. Association for
  Computational Linguistics.

\bibitem[{Liang et~al.(2024)Liang, Liao, Fei, Li, and Jiang}]{liang:2024}
Jinggui Liang, Lizi Liao, Hao Fei, Bobo Li, and Jing Jiang. 2024.
\newblock \href {https://doi.org/10.18653/v1/2024.naacl-long.434} {Actively
  learn from {LLM}s with uncertainty propagation for generalized category
  discovery}.
\newblock In \emph{Proceedings of the 2024 Conference of the North American
  Chapter of the Association for Computational Linguistics: Human Language
  Technologies (Volume 1: Long Papers)}, pages 7845--7858, Mexico City, Mexico.
  Association for Computational Linguistics.

\bibitem[{Liu and Wronski(2018)}]{liu:2018}
Mingnan Liu and Laura Wronski. 2018.
\newblock \href {https://doi.org/10.1177/0894439317695581} {Examining
  completion rates in web surveys via over 25,000 real-world surveys}.
\newblock \emph{Social Science Computer Review}, 36(1):116--124.

\bibitem[{Lowell et~al.(2019)Lowell, Lipton, and
  Wallace}]{lowell-etal-2019-practical}
David Lowell, Zachary~C. Lipton, and Byron~C. Wallace. 2019.
\newblock \href {https://doi.org/10.18653/v1/D19-1003} {Practical obstacles to
  deploying active learning}.
\newblock In \emph{Proceedings of the 2019 Conference on Empirical Methods in
  Natural Language Processing and the 9th International Joint Conference on
  Natural Language Processing (EMNLP-IJCNLP)}, pages 21--30, Hong Kong, China.
  Association for Computational Linguistics.

\bibitem[{Luo et~al.(2023)Luo, Tan, Nguyen, and Du}]{luo:2023}
Haocheng Luo, Wei Tan, Ngoc Nguyen, and Lan Du. 2023.
\newblock \href {https://doi.org/10.18653/v1/2023.findings-emnlp.847}
  {Re-weighting tokens: A simple and effective active learning strategy for
  named entity recognition}.
\newblock In \emph{Findings of the Association for Computational Linguistics:
  EMNLP 2023}, pages 12725--12734, Singapore. Association for Computational
  Linguistics.

\bibitem[{Margatina and Aletras(2023)}]{margatina-aletras-2023-limitations}
Katerina Margatina and Nikolaos Aletras. 2023.
\newblock \href {https://doi.org/10.18653/v1/2023.findings-acl.269} {On the
  limitations of simulating active learning}.
\newblock In \emph{Findings of the Association for Computational Linguistics:
  ACL 2023}, pages 4402--4419, Toronto, Canada. Association for Computational
  Linguistics.

\bibitem[{Margatina et~al.(2022)Margatina, Barrault, and
  Aletras}]{margatina-etal-2022-importance}
Katerina Margatina, Loic Barrault, and Nikolaos Aletras. 2022.
\newblock \href {https://doi.org/10.18653/v1/2022.acl-short.93} {On the
  importance of effectively adapting pretrained language models for active
  learning}.
\newblock In \emph{Proceedings of the 60th Annual Meeting of the Association
  for Computational Linguistics (Volume 2: Short Papers)}, pages 825--836,
  Dublin, Ireland. Association for Computational Linguistics.

\bibitem[{Margatina et~al.(2023)Margatina, Schick, Aletras, and
  Dwivedi-Yu}]{margatina-etal-2023-active}
Katerina Margatina, Timo Schick, Nikolaos Aletras, and Jane Dwivedi-Yu. 2023.
\newblock \href {https://doi.org/10.18653/v1/2023.findings-emnlp.334} {Active
  learning principles for in-context learning with large language models}.
\newblock In \emph{Findings of the Association for Computational Linguistics:
  EMNLP 2023}, pages 5011--5034, Singapore. Association for Computational
  Linguistics.

\bibitem[{Margatina et~al.(2021)Margatina, Vernikos, Barrault, and
  Aletras}]{margatina:2021}
Katerina Margatina, Giorgos Vernikos, Lo{\"i}c Barrault, and Nikolaos Aletras.
  2021.
\newblock \href {https://doi.org/10.18653/v1/2021.emnlp-main.51} {Active
  learning by acquiring contrastive examples}.
\newblock In \emph{Proceedings of the 2021 Conference on Empirical Methods in
  Natural Language Processing}, pages 650--663, Online and Punta Cana,
  Dominican Republic. Association for Computational Linguistics.

\bibitem[{Mendon{\c{c}}a et~al.(2023)Mendon{\c{c}}a, Rei, Coheur, and
  Sardinha}]{mendonca:2023}
V{\^a}nia Mendon{\c{c}}a, Ricardo Rei, Lu{\'i}sa Coheur, and Alberto Sardinha.
  2023.
\newblock \href {https://doi.org/10.1162/coli_a_00473} {Onception: Active
  learning with expert advice for real world machine translation}.
\newblock \emph{Computational Linguistics}, 49(2):325--372.

\bibitem[{Michael et~al.(2023)Michael, Holtzman, Parrish, Mueller, Wang, Chen,
  Madaan, Nangia, Pang, Phang, and Bowman}]{michael:2023}
Julian Michael, Ari Holtzman, Alicia Parrish, Aaron Mueller, Alex Wang,
  Angelica Chen, Divyam Madaan, Nikita Nangia, Richard~Yuanzhe Pang, Jason
  Phang, and Samuel~R. Bowman. 2023.
\newblock \href {https://doi.org/10.18653/v1/2023.acl-long.903} {What do {NLP}
  researchers believe? results of the {NLP} community metasurvey}.
\newblock In \emph{Proceedings of the 61st Annual Meeting of the Association
  for Computational Linguistics (Volume 1: Long Papers)}, pages 16334--16368,
  Toronto, Canada. Association for Computational Linguistics.

\bibitem[{Nachtegael et~al.(2023)Nachtegael, De~Stefani, and
  Lenaerts}]{nachtegael:2023}
Charlotte Nachtegael, Jacopo De~Stefani, and Tom Lenaerts. 2023.
\newblock \href {https://doi.org/10.18653/v1/2023.eacl-demo.14} {{ALAMBIC} :
  Active learning automation methods to battle inefficient curation}.
\newblock In \emph{Proceedings of the 17th Conference of the European Chapter
  of the Association for Computational Linguistics: System Demonstrations},
  pages 117--127, Dubrovnik, Croatia. Association for Computational
  Linguistics.

\bibitem[{Olsson(2009)}]{olsson2009survey}
Fredrik Olsson. 2009.
\newblock \href {https://urn.kb.se/resolve?urn=urn:nbn:se:ri:diva-23510} {A
  literature survey of active machine learning in the context of natural
  language processing}.
\newblock SICS Technical Report T2009:06, Swedish Institute of Computer
  Science.

\bibitem[{Pei et~al.(2022)Pei, Ananthasubramaniam, Wang, Zhou, Dedeloudis,
  Sargent, and Jurgens}]{pei-etal-2022-potato}
Jiaxin Pei, Aparna Ananthasubramaniam, Xingyao Wang, Naitian Zhou, Apostolos
  Dedeloudis, Jackson Sargent, and David Jurgens. 2022.
\newblock \href {https://doi.org/10.18653/v1/2022.emnlp-demos.33} {{POTATO}:
  The portable text annotation tool}.
\newblock In \emph{Proceedings of the 2022 Conference on Empirical Methods in
  Natural Language Processing: System Demonstrations}, pages 327--337, Abu
  Dhabi, UAE. Association for Computational Linguistics.

\bibitem[{Ras et~al.(2022)Ras, Xie, van Gerven, and Doran}]{ras:2022}
Gabrielle Ras, Ning Xie, Marcel van Gerven, and Derek Doran. 2022.
\newblock \href {https://doi.org/10.1613/jair.1.13200} {Explainable deep
  learning: A field guide for the uninitiated}.
\newblock \emph{J. Artif. Int. Res.}, 73.

\bibitem[{Rogers and Luccioni(2024)}]{rogers2024position}
Anna Rogers and Sasha Luccioni. 2024.
\newblock \href {https://openreview.net/forum?id=M2cwkGleRL} {Position: Key
  claims in {LLM} research have a long tail of footnotes}.
\newblock In \emph{Forty-first International Conference on Machine Learning}.

\bibitem[{Romberg and Escher(2022)}]{romberg2022automated}
Julia Romberg and Tobias Escher. 2022.
\newblock \href {https://doi.org/10.1007/978-3-031-15086-9_24} {Automated topic
  categorisation of citizens' contributions: Reducing manual labelling efforts
  through active learning}.
\newblock In \emph{Electronic Government}, pages 369--385, Cham. Springer
  International Publishing.

\bibitem[{Schr{\"o}der et~al.(2022)Schr{\"o}der, Niekler, and
  Potthast}]{schroder-etal-2022-revisiting}
Christopher Schr{\"o}der, Andreas Niekler, and Martin Potthast. 2022.
\newblock \href {https://doi.org/10.18653/v1/2022.findings-acl.172} {Revisiting
  uncertainty-based query strategies for active learning with transformers}.
\newblock In \emph{Findings of the Association for Computational Linguistics:
  ACL 2022}, pages 2194--2203, Dublin, Ireland. Association for Computational
  Linguistics.

\bibitem[{Schröder and Niekler(2020)}]{schroeder2020surveyactivelearningtext}
Christopher Schröder and Andreas Niekler. 2020.
\newblock \href {https://arxiv.org/abs/2008.07267} {A survey of active learning
  for text classification using deep neural networks}.
\newblock \emph{Preprint}, arXiv:2008.07267.

\bibitem[{Shelmanov et~al.(2019)Shelmanov, Liventsev, Kireev, Khromov,
  Panchenko, Fedulova, and Dylov}]{shelmanov:2019}
Artem Shelmanov, Vadim Liventsev, Danil Kireev, Nikita Khromov, Alexander
  Panchenko, Irina Fedulova, and Dmitry~V. Dylov. 2019.
\newblock \href {https://doi.org/10.1109/BIBM47256.2019.8983157} {Active
  learning with deep pre-trained models for sequence tagging of clinical and
  biomedical texts}.
\newblock In \emph{2019 IEEE International Conference on Bioinformatics and
  Biomedicine (BIBM)}, pages 482--489.

\bibitem[{Shelmanov et~al.(2021)Shelmanov, Puzyrev, Kupriyanova, Belyakov,
  Larionov, Khromov, Kozlova, Artemova, Dylov, and Panchenko}]{shelmanov:2021}
Artem Shelmanov, Dmitri Puzyrev, Lyubov Kupriyanova, Denis Belyakov, Daniil
  Larionov, Nikita Khromov, Olga Kozlova, Ekaterina Artemova, Dmitry~V. Dylov,
  and Alexander Panchenko. 2021.
\newblock \href {https://doi.org/10.18653/v1/2021.eacl-main.145} {Active
  learning for sequence tagging with deep pre-trained models and {B}ayesian
  uncertainty estimates}.
\newblock In \emph{Proceedings of the 16th Conference of the European Chapter
  of the Association for Computational Linguistics: Main Volume}, pages
  1698--1712, Online. Association for Computational Linguistics.

\bibitem[{Shen et~al.(2017)Shen, Yun, Lipton, Kronrod, and
  Anandkumar}]{shen-etal-2017-deep}
Yanyao Shen, Hyokun Yun, Zachary Lipton, Yakov Kronrod, and Animashree
  Anandkumar. 2017.
\newblock \href {https://doi.org/10.18653/v1/W17-2630} {Deep active learning
  for named entity recognition}.
\newblock In \emph{Proceedings of the 2nd Workshop on Representation Learning
  for {NLP}}, pages 252--256, Vancouver, Canada. Association for Computational
  Linguistics.

\bibitem[{Shnarch et~al.(2022{\natexlab{a}})Shnarch, Halfon, Gera, Danilevsky,
  Katsis, Choshen, Santillan~Cooper, Epelboim, Zhang, and Wang}]{shnarch:2022}
Eyal Shnarch, Alon Halfon, Ariel Gera, Marina Danilevsky, Yannis Katsis, Leshem
  Choshen, Martin Santillan~Cooper, Dina Epelboim, Zheng Zhang, and Dakuo Wang.
  2022{\natexlab{a}}.
\newblock \href {https://doi.org/10.18653/v1/2022.emnlp-demos.16} {Label
  sleuth: From unlabeled text to a classifier in a few hours}.
\newblock In \emph{Proceedings of the 2022 Conference on Empirical Methods in
  Natural Language Processing: System Demonstrations}, pages 159--168, Abu
  Dhabi, UAE. Association for Computational Linguistics.

\bibitem[{Shnarch et~al.(2022{\natexlab{b}})Shnarch, Halfon, Gera, Danilevsky,
  Katsis, Choshen, Santillan~Cooper, Epelboim, Zhang, Wang, Yip, Ein-Dor,
  Dankin, Shnayderman, Aharonov, Li, Liberman, Levin~Slesarev, Newton,
  Ofek-Koifman, Slonim, and Katz}]{shnarch-etal-2022-label}
Eyal Shnarch, Alon Halfon, Ariel Gera, Marina Danilevsky, Yannis Katsis, Leshem
  Choshen, Martin Santillan~Cooper, Dina Epelboim, Zheng Zhang, Dakuo Wang,
  Lucy Yip, Liat Ein-Dor, Lena Dankin, Ilya Shnayderman, Ranit Aharonov, Yunyao
  Li, Naftali Liberman, Philip Levin~Slesarev, Gwilym Newton, and 3 others.
  2022{\natexlab{b}}.
\newblock \href {https://doi.org/10.18653/v1/2022.emnlp-demos.16} {Label
  sleuth: From unlabeled text to a classifier in a few hours}.
\newblock In \emph{Proceedings of the 2022 Conference on Empirical Methods in
  Natural Language Processing: System Demonstrations}, pages 159--168, Abu
  Dhabi, UAE. Association for Computational Linguistics.

\bibitem[{Subramonian et~al.(2023)Subramonian, Yuan, Daum{\'e}~III, and
  Blodgett}]{subramonian:2023}
Arjun Subramonian, Xingdi Yuan, Hal Daum{\'e}~III, and Su~Lin Blodgett. 2023.
\newblock \href {https://doi.org/10.18653/v1/2023.findings-acl.202} {It takes
  two to tango: Navigating conceptualizations of {NLP} tasks and measurements
  of performance}.
\newblock In \emph{Findings of the Association for Computational Linguistics:
  ACL 2023}, pages 3234--3279, Toronto, Canada. Association for Computational
  Linguistics.

\bibitem[{Tan et~al.(2024)Tan, Li, Wang, Beigi, Jiang, Bhattacharjee, Karami,
  Li, Cheng, and Liu}]{tan-etal-2024-large}
Zhen Tan, Dawei Li, Song Wang, Alimohammad Beigi, Bohan Jiang, Amrita
  Bhattacharjee, Mansooreh Karami, Jundong Li, Lu~Cheng, and Huan Liu. 2024.
\newblock \href {https://doi.org/10.18653/v1/2024.emnlp-main.54} {Large
  language models for data annotation and synthesis: A survey}.
\newblock In \emph{Proceedings of the 2024 Conference on Empirical Methods in
  Natural Language Processing}, pages 930--957, Miami, Florida, USA.
  Association for Computational Linguistics.

\bibitem[{Tomanek and Olsson(2009)}]{tomanek-olsson-2009-web}
Katrin Tomanek and Fredrik Olsson. 2009.
\newblock \href {https://aclanthology.org/W09-1906} {A web survey on the use of
  active learning to support annotation of text data}.
\newblock In \emph{Proceedings of the {NAACL} {HLT} 2009 Workshop on Active
  Learning for Natural Language Processing}, pages 45--48, Boulder, Colorado.
  Association for Computational Linguistics.

\bibitem[{Tonneau et~al.(2022)Tonneau, Adjodah, Palotti, Grinberg, and
  Fraiberger}]{tonneau:2022}
Manuel Tonneau, Dhaval Adjodah, Joao Palotti, Nir Grinberg, and Samuel
  Fraiberger. 2022.
\newblock \href {https://doi.org/10.18653/v1/2022.acl-long.453} {Multilingual
  detection of personal employment status on {T}witter}.
\newblock In \emph{Proceedings of the 60th Annual Meeting of the Association
  for Computational Linguistics (Volume 1: Long Papers)}, pages 6564--6587,
  Dublin, Ireland. Association for Computational Linguistics.

\bibitem[{Tsvigun et~al.(2022{\natexlab{a}})Tsvigun, Lysenko, Sedashov,
  Lazichny, Damirov, Karlov, Belousov, Sanochkin, Panov, Panchenko, Burtsev,
  and Shelmanov}]{tsvigun:2022b}
Akim Tsvigun, Ivan Lysenko, Danila Sedashov, Ivan Lazichny, Eldar Damirov,
  Vladimir Karlov, Artemy Belousov, Leonid Sanochkin, Maxim Panov, Alexander
  Panchenko, Mikhail Burtsev, and Artem Shelmanov. 2022{\natexlab{a}}.
\newblock \href {https://doi.org/10.18653/v1/2022.findings-emnlp.377} {Active
  learning for abstractive text summarization}.
\newblock In \emph{Findings of the Association for Computational Linguistics:
  EMNLP 2022}, pages 5128--5152, Abu Dhabi, United Arab Emirates. Association
  for Computational Linguistics.

\bibitem[{Tsvigun et~al.(2022{\natexlab{b}})Tsvigun, Sanochkin, Larionov,
  Kuzmin, Vazhentsev, Lazichny, Khromov, Kireev, Rubashevskii, Shahmatova,
  Dylov, Galitskiy, and Shelmanov}]{tsvigun:2022}
Akim Tsvigun, Leonid Sanochkin, Daniil Larionov, Gleb Kuzmin, Artem Vazhentsev,
  Ivan Lazichny, Nikita Khromov, Danil Kireev, Aleksandr Rubashevskii, Olga
  Shahmatova, Dmitry~V. Dylov, Igor Galitskiy, and Artem Shelmanov.
  2022{\natexlab{b}}.
\newblock \href {https://doi.org/10.18653/v1/2022.emnlp-demos.41} {{ALT}oolbox:
  A set of tools for active learning annotation of natural language texts}.
\newblock In \emph{Proceedings of the 2022 Conference on Empirical Methods in
  Natural Language Processing: System Demonstrations}, pages 406--434, Abu
  Dhabi, UAE. Association for Computational Linguistics.

\bibitem[{Tsvigun et~al.(2022{\natexlab{c}})Tsvigun, Shelmanov, Kuzmin,
  Sanochkin, Larionov, Gusev, Avetisian, and
  Zhukov}]{tsvigun-etal-2022-towards}
Akim Tsvigun, Artem Shelmanov, Gleb Kuzmin, Leonid Sanochkin, Daniil Larionov,
  Gleb Gusev, Manvel Avetisian, and Leonid Zhukov. 2022{\natexlab{c}}.
\newblock \href {https://doi.org/10.18653/v1/2022.findings-naacl.90} {Towards
  computationally feasible deep active learning}.
\newblock In \emph{Findings of the Association for Computational Linguistics:
  NAACL 2022}, pages 1198--1218, Seattle, United States. Association for
  Computational Linguistics.

\bibitem[{Van Der~Meer et~al.(2024)Van Der~Meer, Falk, Murukannaiah, and
  Liscio}]{van-der-meer-etal-2024-annotator}
Michiel Van Der~Meer, Neele Falk, Pradeep~K. Murukannaiah, and Enrico Liscio.
  2024.
\newblock \href {https://doi.org/10.18653/v1/2024.emnlp-main.1031}
  {Annotator-centric active learning for subjective {NLP} tasks}.
\newblock In \emph{Proceedings of the 2024 Conference on Empirical Methods in
  Natural Language Processing}, pages 18537--18555, Miami, Florida, USA.
  Association for Computational Linguistics.

\bibitem[{Wang and Plank(2023)}]{wang-plank-2023-actor}
Xinpeng Wang and Barbara Plank. 2023.
\newblock \href {https://doi.org/10.18653/v1/2023.emnlp-main.126} {{ACTOR}:
  Active learning with annotator-specific classification heads to embrace human
  label variation}.
\newblock In \emph{Proceedings of the 2023 Conference on Empirical Methods in
  Natural Language Processing}, pages 2046--2052, Singapore. Association for
  Computational Linguistics.

\bibitem[{Weber and Plank(2023)}]{weber:2023}
Leon Weber and Barbara Plank. 2023.
\newblock \href {https://doi.org/10.18653/v1/2023.findings-acl.562}
  {{A}ctive{AED}: A human in the loop improves annotation error detection}.
\newblock In \emph{Findings of the Association for Computational Linguistics:
  ACL 2023}, pages 8834--8845, Toronto, Canada. Association for Computational
  Linguistics.

\bibitem[{Wiechmann et~al.(2021)Wiechmann, Yimam, and
  Biemann}]{wiechmann-etal-2021-activeanno}
Max Wiechmann, Seid~Muhie Yimam, and Chris Biemann. 2021.
\newblock \href {https://doi.org/10.18653/v1/2021.naacl-demos.12}
  {{A}ctive{A}nno: General-purpose document-level annotation tool with active
  learning integration}.
\newblock In \emph{Proceedings of the 2021 Conference of the North American
  Chapter of the Association for Computational Linguistics: Human Language
  Technologies: Demonstrations}, pages 99--105, Online. Association for
  Computational Linguistics.

\bibitem[{Xia et~al.(2024)Xia, Liu, Yu, Kim, Rossi, Rao, Mai, and
  Li}]{xia:2024}
Yu~Xia, Xu~Liu, Tong Yu, Sungchul Kim, Ryan Rossi, Anup Rao, Tung Mai, and
  Shuai Li. 2024.
\newblock \href {https://doi.org/10.18653/v1/2024.naacl-long.479}
  {Hallucination diversity-aware active learning for text summarization}.
\newblock In \emph{Proceedings of the 2024 Conference of the North American
  Chapter of the Association for Computational Linguistics: Human Language
  Technologies (Volume 1: Long Papers)}, pages 8665--8677, Mexico City, Mexico.
  Association for Computational Linguistics.

\bibitem[{Xia et~al.(2025)Xia, Mukherjee, Xie, Wu, Li, Aponte, Lyu, Barrow,
  Chen, Dernoncourt, Kveton, Yu, Zhang, Gu, Ahmed, Wang, Chen, Deilamsalehy,
  Kim, Hu, Zhao, Lipka, Yoon, Huang, Wang, Mathur, Pal, Mukherjee, Zhang, Park,
  Nguyen, Luo, Rossi, and McAuley}]{xia-etal-2025-selection}
Yu~Xia, Subhojyoti Mukherjee, Zhouhang Xie, Junda Wu, Xintong Li, Ryan Aponte,
  Hanjia Lyu, Joe Barrow, Hongjie Chen, Franck Dernoncourt, Branislav Kveton,
  Tong Yu, Ruiyi Zhang, Jiuxiang Gu, Nesreen~K. Ahmed, Yu~Wang, Xiang Chen,
  Hanieh Deilamsalehy, Sungchul Kim, and 15 others. 2025.
\newblock \href {https://doi.org/10.18653/v1/2025.acl-long.708} {From selection
  to generation: A survey of {LLM}-based active learning}.
\newblock In \emph{Proceedings of the 63rd Annual Meeting of the Association
  for Computational Linguistics (Volume 1: Long Papers)}, pages 14552--14569,
  Vienna, Austria. Association for Computational Linguistics.

\bibitem[{Xiao et~al.(2023)Xiao, Dong, Zhao, Wu, Lin, Chen, and
  Wang}]{xiao:2023}
Ruixuan Xiao, Yiwen Dong, Junbo Zhao, Runze Wu, Minmin Lin, Gang Chen, and
  Haobo Wang. 2023.
\newblock \href {https://doi.org/10.18653/v1/2023.emnlp-main.896} {{F}ree{AL}:
  Towards human-free active learning in the era of large language models}.
\newblock In \emph{Proceedings of the 2023 Conference on Empirical Methods in
  Natural Language Processing}, pages 14520--14535, Singapore. Association for
  Computational Linguistics.

\bibitem[{Yuan et~al.(2020)Yuan, Lin, and Boyd-Graber}]{yuan:2020}
Michelle Yuan, Hsuan-Tien Lin, and Jordan Boyd-Graber. 2020.
\newblock \href {https://doi.org/10.18653/v1/2020.emnlp-main.637} {Cold-start
  active learning through self-supervised language modeling}.
\newblock In \emph{Proceedings of the 2020 Conference on Empirical Methods in
  Natural Language Processing (EMNLP)}, pages 7935--7948, Online. Association
  for Computational Linguistics.

\bibitem[{Y{\"u}ksel et~al.(2023)Y{\"u}ksel, Gunduz, Al-badrashiny, and
  Sawaf}]{yuksel:2023}
Kamer Y{\"u}ksel, Ahmet Gunduz, Mohamed Al-badrashiny, and Hassan Sawaf. 2023.
\newblock \href {https://doi.org/10.18653/v1/2023.acl-industry.33}
  {{E}volve{MT}: an ensemble {MT} engine improving itself with usage only}.
\newblock In \emph{Proceedings of the 61st Annual Meeting of the Association
  for Computational Linguistics (Volume 5: Industry Track)}, pages 341--346,
  Toronto, Canada. Association for Computational Linguistics.

\bibitem[{Zeng et~al.(2019)Zeng, Garg, Chatterjee, Nallasamy, and
  Paulik}]{zeng:2019}
Xiangkai Zeng, Sarthak Garg, Rajen Chatterjee, Udhyakumar Nallasamy, and
  Matthias Paulik. 2019.
\newblock \href {https://doi.org/10.18653/v1/D19-6110} {Empirical evaluation of
  active learning techniques for neural {MT}}.
\newblock In \emph{Proceedings of the 2nd Workshop on Deep Learning Approaches
  for Low-Resource NLP (DeepLo 2019)}, pages 84--93, Hong Kong, China.
  Association for Computational Linguistics.

\bibitem[{Zhang et~al.(2023)Zhang, Li, Ma, Zhou, and Zou}]{zhang:2023}
Ruoyu Zhang, Yanzeng Li, Yongliang Ma, Ming Zhou, and Lei Zou. 2023.
\newblock \href {https://doi.org/10.18653/v1/2023.findings-emnlp.872}
  {{LLM}a{AA}: Making large language models as active annotators}.
\newblock In \emph{Findings of the Association for Computational Linguistics:
  EMNLP 2023}, pages 13088--13103, Singapore. Association for Computational
  Linguistics.

\bibitem[{Zhang et~al.(2022)Zhang, Strubell, and Hovy}]{zhang-etal-2022-survey}
Zhisong Zhang, Emma Strubell, and Eduard Hovy. 2022.
\newblock \href {https://doi.org/10.18653/v1/2022.emnlp-main.414} {A survey of
  active learning for natural language processing}.
\newblock In \emph{Proceedings of the 2022 Conference on Empirical Methods in
  Natural Language Processing}, pages 6166--6190, Abu Dhabi, United Arab
  Emirates. Association for Computational Linguistics.

\bibitem[{Zhao et~al.(2020)Zhao, Zhang, Zhou, and Zhang}]{zhao:2020}
Yuekai Zhao, Haoran Zhang, Shuchang Zhou, and Zhihua Zhang. 2020.
\newblock \href {https://doi.org/10.18653/v1/2020.findings-emnlp.162} {Active
  learning approaches to enhancing neural machine translation}.
\newblock In \emph{Findings of the Association for Computational Linguistics:
  EMNLP 2020}, pages 1796--1806, Online. Association for Computational
  Linguistics.

\bibitem[{Zhou et~al.(2022)Zhou, Blodgett, Trischler, Daum{\'e}~III, Suleman,
  and Olteanu}]{zhou:2022}
Kaitlyn Zhou, Su~Lin Blodgett, Adam Trischler, Hal Daum{\'e}~III, Kaheer
  Suleman, and Alexandra Olteanu. 2022.
\newblock \href {https://doi.org/10.18653/v1/2022.naacl-main.24}
  {Deconstructing {NLG} evaluation: Evaluation practices, assumptions, and
  their implications}.
\newblock In \emph{Proceedings of the 2022 Conference of the North American
  Chapter of the Association for Computational Linguistics: Human Language
  Technologies}, pages 314--324, Seattle, United States. Association for
  Computational Linguistics.

\end{thebibliography}
